\pdfoutput=1
\documentclass{article}
\usepackage{iclr2027_conference,times}
\usepackage[T1,OT1]{fontenc}
\DeclareTextAccentDefault{\k}{T1}

\usepackage{amsmath,amsfonts,bm}

\def\eqref#1{equation~\ref{#1}}

\def\1{\bm{1}}

\DeclareMathAlphabet{\mathsfit}{\encodingdefault}{\sfdefault}{m}{sl}
\SetMathAlphabet{\mathsfit}{bold}{\encodingdefault}{\sfdefault}{bx}{n}

\usepackage{hyperref}
\usepackage{url}

\usepackage{graphicx}
\usepackage{booktabs}
\usepackage{colortbl}
\usepackage{amssymb}
\usepackage{multirow}
\usepackage{float}
\title{BellowsKV: One Latent for Many Layers, \\ Expanded on Demand}

\iclrfinalcopy

\author{%
  Jiamu Zhang\quad
  Hanjie Chen\\[3pt]
  Rice University\\
  \texttt{mz81@rice.edu}\quad\texttt{hanjie@rice.edu}
}

\begin{document}

\maketitle
\lhead{Preprint}

\begin{abstract}
     Large language models (LLMs) are now widely utilized as the backbone of agents that reason over many turns, call tools, and pursue long-horizon tasks. To serve such a context without recomputing the past, a model stores the attention keys and values of all past tokens in a key-value (KV) cache. Each layer maintains its own cache and re-reads it at every generation step, so KV-cache memory can become the main bottleneck for GPU serving. Most compression methods operate within a single layer, by quantizing the cache entries, reducing the channel dimension, or evicting less important tokens from the cache. All three lose substantial accuracy at high compression ratios. A more recent line of work shares one compressed cache across adjacent layers to go beyond this limit. However, existing cross-layer methods also degrade sharply at high compression ratios. We find that the representation of KV Cache between adjacent layers are quite different, whereas the input hidden states are highly similar. We also find that, under the same memory budget, reducing the dimension of this shared latent loses far more accuracy than quantizing it. Based on these findings, we propose BellowsKV, a training-free method that maps the hidden states of n adjacent layers to one shared latent per token. At inference time, each layer \textit{read} its own keys and values from this latent through a layer-specific projection. We keep the latent at full dimension and compress it in the two other ways instead: BellowsKV-Quant stores the latent at low precision, which is nearly lossless because the latent has no dominant outlier channels; BellowsKV-Evict makes one eviction decision per block, which the layers can share because they rank token importance very similarly. Across Llama-3.1-8B, Mistral-7B and Qwen2.5-7B, BellowsKV-Quant retains about 99\% of full-cache accuracy at 3.6$\times$ compression and 97-98\% at 13-14$\times$, where the strongest existing cross-layer method retains at most 84\% at 16$\times$. Code: \url{https://github.com/MorrisZJ/BellowsKV}

\end{abstract}

\section{Introduction}

The way large language models are used has shifted from answering questions to carrying out tasks. Agents
debug a repository over hundreds of tool calls with the accumulated history in view \citep{yang2024sweagentagentcomputerinterfacesenable, jimenez2024swebenchlanguagemodelsresolve},
assistants read and cross-reference whole document collections in one pass \citep{bai2024longbenchbilingualmultitaskbenchmark, hsieh2024rulerwhatsrealcontext}, and
reasoning models think in traces tens of thousands of tokens long before they answer \citep{openai2026openaio1card, Guo_2025}.
Context windows have chased these workloads, from the 4K to 128K as the open-weight default
\citep{grattafiori2024llama3herdmodels} and past a million tokens at the frontier \citep{geminiteam2024gemini15unlockingmultimodal, yang2025qwen251mtechnicalreport}. 
The key-value (KV) cache holds the attention keys and values of every token in the window, and each layer keeps its own cache. This way, we do not need to recompute past tokens, because we can read the cache in full at every decoding step.
At this scale the cache stops being a rounding error on the weights and becomes their rival: on
Llama-3.1-8B a single 131K-token sequence carries a 16.8 GB cache in 16-bit precision, slightly more than
the model itself, after grouped-query attention has already shrunk it fourfold \citep{ainslie2023gqatraininggeneralizedmultiquery}. 
In the real system, the LLM weights are loaded once and shared by every request in a batch, but each sequence needs its own KV cache. Serving is economical only when many sequences run at the same time (Kwon et al., 2023), so the total cache size grows with the batch size while the weights stay fixed. A batch of eight 131K-token sequences, for example, holds 134 GB of cache against 16 GB of weights. Compressing it by 4x, where most published methods operate, would still leave 33 GB. This paper targets an order of magnitude instead.

\begin{figure}[t]
  \centering
  \includegraphics[width=0.98\linewidth]{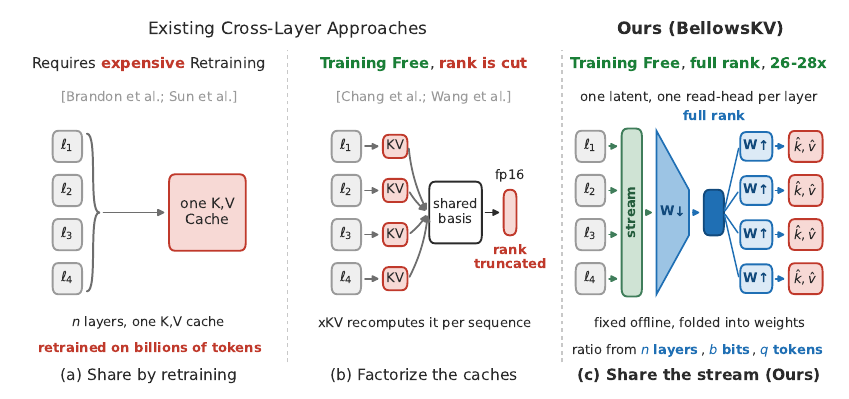}
    \caption{\textbf{Comparison of different cross-layer KV-Cache compression methods}: (a) Trained-to-share architectures reach one cache only by retraining.
    (b) Training-free methods factorize the caches
    (c) BellowsKV folds the block's hidden states into one latent, buying the compression ratio from the numbers of shared layers, bits and tokens instead.}
  \label{fig:motivation}
\end{figure}

Nearly all existing KV cache compression methods work inside a single layer. Cached keys and values are quantized to fewer bits \citep{liu2024kivi, su2025rotatekvaccuraterobust2bit, zandieh2025turboquantonlinevectorquantization, han2025polarquantquantizingkvcaches}, tokens judged unimportant are evicted
\citep{zhang2023h2oheavyhitteroracleefficient, li2024snapkvllmknowslooking, xiao2024streamingllm}, and feature channels are pruned or projected to lower rank
\citep{chang2024palucompressingkvcachelowrank, saxena2024eigenattentionattentionlowrank, staniszewski2026kvcachetransformcoding}. 
All three directions work at moderate compression ratios, but each of them is close to its limit when further compressing the cache. 

Quantization loses accuracy sharply below two bits, and accuracy depends on extra machinery around the quantization methods, such as choosing the axis along which numbers share a scale, rotating the cache so that no channel dominates a scale, or keeping a sparse set of outliers in full precision \citep{hooper2024kvquant}. Token eviction cannot be undone when regret, because a token dropped after prefill is gone for every later decoding step and for every follow-up turn, even if attention would have needed it. Reducing the channel dimension removes components of the keys and values that attention still uses. A recent position paper argues that at a matched memory budget, quantization is preferable to rank reduction \citep{salfati2026quantizationdominatesrankreduction}.

However, adjacent layers compute their caches from almost the same hidden states, yet every layer still stores its own cache in full. Sharing a cache across layers
is therefore the natural next move, and it has recently been tried. 
xKV \citep{chang2026xkvcrosslayerkvcachecompression} factorizes the caches of a window of layers into one shared low-rank basis, recomputed for every sequence at prefill. CommonKV \citep{wang2025commonkvcompressingkvcache} instead derives a shared projection from the weight matrices of the grouped layers. Both work at 4x compression, but at 16x, xKV retains at most 84\% of full-cache accuracy and CommonKV at most 54\%. To our knowledge, no prior work explains why cross-layer methods degrade at these ratios, or what a block of layers needs in order to share one cache without losing accuracy.
Appendix~\ref{app:related} reviews all three families in full, together with the architectures that are trained to share a cache.

Therefore, we propose BellowsKV, a training-free KV cache compression method built on these three measurements. 
For each block of n adjacent layers, we fit a single projection offline, and it maps the hidden states of the block to one latent vector per token, and each layer reconstructs its own keys and values from that latent through its own up-projection. Two operators then compress the latent: BellowsKV-Quant stores it with uniform low-bit quantization, and BellowsKV-Evict drops tokens with a single eviction decision for the whole block. Our contributions are:

\begin{itemize}
    \item We study what a block of adjacent layers can share. The keys and values of adjacent layers
    are nearly orthogonal, while the hidden states they are computed from are highly similar. The
    representation shared by a block tolerates quantization and eviction, but not dimension reduction
    (Section~\ref{sec:motivation}).
    \item We propose BellowsKV, a training-free method that stores one latent per block of $n$ layers,
    together with two operators that compress this latent, BellowsKV-Quant and BellowsKV-Evict
    (Section~\ref{sec:method}).
\end{itemize}

\section{Motivation: What Should a Block of Layers Store?}
\label{sec:motivation}

\begin{figure}[ht]
  \centering
  \includegraphics[width=0.85\linewidth]{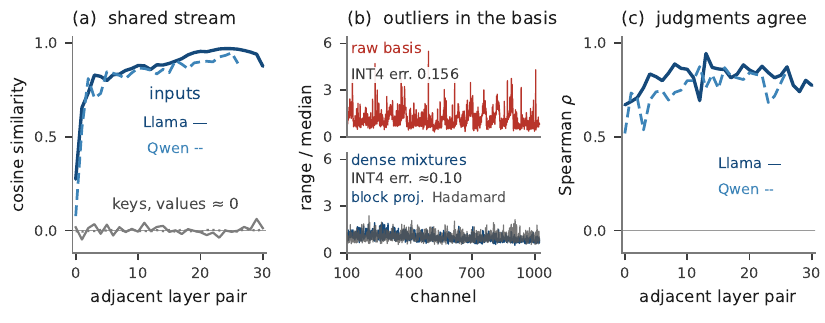}
    \caption{(a) Cosine similarity between adjacent layers. The keys and values are close
    to zero, while the hidden states reach $0.87$ on Llama and $0.82$ on Qwen. (b) Per-channel dynamic
    range of the keys. The raw cache concentrates the range in a few
    channels (top). A data-free Hadamard rotation and our block-level projection are both flat
    (bottom), and the relative INT4 error falls from $0.156$ to about $0.10$. (c) Spearman correlation
    between the token importance scores of adjacent layers, between $0.76$ and $0.80$.}
  \label{fig:motivation}
\end{figure}

Every cache compression method makes two decisions: what to store, and how coarsely to store it.
Existing methods only make the second: quantization, eviction and channel reduction keep one
private cache per layer, as the architecture defines it, and compress that cache in place. This
paper starts from the first decision. We ask what a block of $n$ layers should store, under two
requirements: every layer must still recover its own keys and values from the stored object, and
the object must remain accurate when stored coarsely. Figure~\ref{fig:motivation} answers with
measurements on the uncompressed models, which do not assume any particular compression method.

\vspace{-2mm}

\subsection{Adjacent layers have similar hidden states but dissimilar cache}
\label{sec:ob1}

A natural reading of cross-layer redundancy is that a later layer could reuse an earlier layer's cache. Our measurements rule this out. Across adjacent layer pairs of both models, the per-token cosine similarity between neighboring keys is close to zero, and the same holds for values (Figure~\ref{fig:motivation}a). Substituting one shared key across even two layers increases perplexity significantly, even when the shared key is chosen by an oracle criteria (Appendix~\ref{app:negatives}).

The reason is that adjacent layers compute their keys and values from the same input, but with different projection matrices. Therefore, the keys and values of two adjacent layers are not similar to each other. In contrast, the hidden states remain similar across adjacent layers, and the similarity increases with depth (Figure~\ref{fig:motivation}a), and the same observation is also reported by \citet{wang2025commonkvcompressingkvcache}, which attribute such observation to the residual connection. These results show that the redundancy between adjacent layers exists in the inputs of the caches, and not in the caches themselves. Several architectures are trained to share one cache across a group of layers \citep{brandon2024reducingtransformerkeyvaluecache, sun2024cacheoncedecoderdecoderarchitectures, zuhri2024mlkvmultilayerkeyvalueheads, wu2024layercondensedkvcacheefficient}. These architectures show that a group of layers can work with a single cache. However, they require training the attention layers on billions of tokens, which is not possible for a model that is already trained. \emph{Therefore, a block of layers should store one representation computed from its hidden states, rather than from its caches.} Each layer then reconstructs its own keys and values from this representation through its own up-projection.

\subsection{What aggregation removes, and the one thing it cannot}
\label{sec:ob2}

The requirement above fixes the structure of the shared representation. To serve $n$ layers at once, the representation must combine the channels of all $n$ layers into common coordinates, and it must store one entry per token for the whole block. We now ask how coarsely such a representation can be stored. We consider three forms of compression: quantization (fewer bits per number), eviction (fewer tokens), and dimension reduction (fewer coordinates per token).

\vspace{-2mm}

\paragraph{Observation 1: The shared representation has no dominant outlier channels}\label{par:amplitude}
Uniform quantization fails when a few channels have a much larger dynamic range than the rest of the channels because these channels determine the scale of the group. 
The KV cache is known to contain such channels \citep{liu2024kivi, hooper2024kvquant};
see Figure~\ref{fig:motivation}b (top). A common solution is to rotate the cache before
quantization, so that every stored coordinate becomes a mixture of all channels and no single
channel determines the scale. However, the rotation must be applied again at every read
\citep{ashkboos2024quarotoutlierfree4bitinference, liu2025spinquantllmquantizationlearned,
su2025rotatekvaccuraterobust2bit, han2025polarquantquantizingkvcaches,
zhou2026oscarofflinespectralcovarianceaware, su2026oscaroccamsrazorextreme}. The shared representation already has this property, because every coordinate it stores is a linear
combination of channels from all $n$ layers. To verify this, we construct the block-level
representation defined above and quantize it with a uniform quantization. Also, we apply the same uniform quantization to a rotated copy of the raw cache. The two are equally flat, and the quantization error of INT4 precision format is $0.10$ for the shared representation against $0.156$ for the raw cache
(Figure~\ref{fig:motivation}b, bottom). Therefore, the shared representation does not need a
rotation before it is stored.

\vspace{-2mm}

\paragraph{Observation 2: The layers of a block agree on which tokens are important}\label{par:judgment}
A token can be evicted only if it is unimportant, and each layer decides this from its own attention scores. 
We therefore compare the importance scores that the layers of a block assign to the same tokens. Adjacent layers rank tokens
very similarly, with a Spearman correlation between $0.76$ and $0.80$, and the correlation
increases with depth (Figure~\ref{fig:motivation}c). The same agreement between neighboring layers
is also reported in previous work \citep{wu2025hshare, liu2024minicachekvcachecompression, ma2025compressingkvcachelongcontext}. A single
eviction decision, computed from the scores of all $n$ layers, can therefore replace the $n$
per-layer decisions.

\vspace{-2mm}

\paragraph{Observation 3: The dimension of the shared representation cannot be reduced without a large loss of accuracy}\label{par:dimension}
The first two properties follow from mixing the channels of $n$ layers. However, mixing does not reduce
the rank of the representation. Therefore, storing fewer coordinates per token removes information
that the layers still need at read time. We do not measure this property in this section. We test
it directly in Section~\ref{sec:exp_rank}, where we compare two ways to reach the same memory
budget: reducing the dimension of the representation, and reducing the number of bits per
coordinate.

These three observations define the design of BellowsKV.\footnote{The name refers to a bellows,
which folds a large volume into a small one and expands it again on demand.} For each block of $n$
layers, we compute one shared representation in closed form from the activation statistics of the
model, and each layer reconstructs its own keys and values from this representation when it reads
the cache. Observation~1 and Observation~2 lead to the two
operators of BellowsKV: BellowsKV-Quant stores the representation with uniform low-bit
quantization, and BellowsKV-Evict applies one eviction decision to the whole block
(Section~\ref{sec:method}). 

\vspace{-2mm}

\section{Method}
\label{sec:method}

A decoder-only transformer has $L$ layers of attention modules with hidden width $d$. At layer $\ell$ and position $t$ in a sequence of token,
attention module reads the hidden state $\mathbf{h}_{\ell,t}\in\mathbb{R}^{d}$ and forms
$\mathbf{k}_{\ell,t}=\mathbf{W}_K^{\ell\,\top}\mathbf{h}_{\ell,t}$ and
$\mathbf{v}_{\ell,t}=\mathbf{W}_V^{\ell\,\top}\mathbf{h}_{\ell,t}$ with projection matrices
$\mathbf{W}_K^{\ell},\mathbf{W}_V^{\ell}\in\mathbb{R}^{d\times p}$, both cached for every layer
and every past position, and for simplicity we write attention as single-headed with key width $p$. Since the construction never looks inside the
key vector, the MHA and GQA \citep{ainslie2023gqatraininggeneralizedmultiquery} follow verbatim. Bold lowercase letters are column vectors and bold uppercase letters are matrices;
$[\mathbf{a};\mathbf{b}]$ stacks $\mathbf{a}$ on top of $\mathbf{b}$, and $\|\cdot\|_F$ is the
Frobenius norm. Bias terms are present in every computation but omitted from the notation.

The $L$ layers are partitioned into consecutive, non-overlapping blocks of $n$ layers,
$\mathcal{B}=(\ell_i,\ell_{i+1},\dots,\ell_{i+n-1})$; for example, $n{=}3$ gives the blocks
$(\ell_0,\ell_1,\ell_2)$, $(\ell_3,\ell_4,\ell_5)$, and so on, and a final block shorter than
$n$ is kept as is. 

The method has two parts. The first part is a projection that is computed once per model, offline, from the activation statistics of the block, and is then held fixed for every sequence. 
This projection maps the hidden states of the block to one latent vector per token, which we call the \emph{consensus}, and it produces all of the compression. 
The second part is a low-rank correction, which we call the \emph{whisper}, and it is fitted at prefill
for each sequence.
In the following section, we write the key side only: values have their own latent and up-projections, constructed identically
with $\mathbf{W}_V$ in place of $\mathbf{W}_K$, and we mark $K$ and $V$ explicitly only where the
two appear together or are treated differently.

\subsection{One latent per block, one up-projection per layer}
\label{sec:object}

For a block $\mathcal{B}$ and token $t$, BellowsKV replaces the block's $n$ cached keys and values with two latents $\mathbf{c}^{K}_t,\mathbf{c}^{V}_t\in\mathbb{R}^{R}$ ($R\ll np$), from which each layer recovers its own key through a fixed up-projection. For example:
\begin{equation}
\mathbf{c}_t=\mathbf{W}_{\downarrow}^{\top}\,\mathbf{h}_{\mathcal{B},t},\qquad
\hat{\mathbf{k}}_{\ell_i,t}=\mathbf{W}_{\uparrow}^{(i)\,\top}\,\mathbf{c}_t,\qquad
\mathbf{h}_{\mathcal{B},t}=\big[\mathbf{h}_{\ell_1,t};\dots;\mathbf{h}_{\ell_n,t}\big]\in\mathbb{R}^{nd},
\label{eq:latent}
\end{equation}
where $i\in\{1,\dots,n\}$ indexes the layers inside the block. The down-projection $\mathbf{W}_{\downarrow}\in\mathbb{R}^{nd\times R}$
is shared by the whole block and applied to the vertical stack $\mathbf{h}_{\mathcal{B},t}$ of
its $n$ hidden states. The up-projection $\mathbf{W}_{\uparrow}^{(i)}\in\mathbb{R}^{R\times p}$ belongs to layer $\ell_i$.
It maps the shared latent back to $\hat{\mathbf{k}}_{\ell_i,t}$, which is an approximation of the
exact key $\mathbf{k}_{\ell_i,t}=\mathbf{W}_K^{\ell_i\,\top}\mathbf{h}_{\ell_i,t}$ of the
uncompressed model. All $n$ layers read the same latent $\mathbf{c}_t$, and each layer applies its
own up-projection. The values use an independent latent of the same form,
$\mathbf{c}^{V}_t=\mathbf{W}_{\downarrow}^{V\,\top}\mathbf{h}_{\mathcal{B},t}$, with its own
up-projections $\mathbf{W}_{\uparrow}^{V,(i)}$. The construction is identical.

The cache therefore holds, per block and per token, the pair
$(\mathbf{c}^{K}_t,\mathbf{c}^{V}_t)$ of $2R$ numbers in place of $2np$, plus a short per-token
whisper correction (Section~\ref{sec:decode}). All four factor sets are fixed model parameters,
computed once offline (Section~\ref{sec:construction}) and folded into the weights, so the
compression itself is never recomputed per sequence; the whisper is the one per-sequence
addition, and it is small. This is exactly the object Section~\ref{sec:motivation} argued a
block should store.

\subsection{Calibrating the projection and the up-projections}
\label{sec:construction}

To minimize the reconstruction error between the exact cache and the cache BellowsKV reads back
from the latent, we fit the factors on calibration data. To set this up, we stack a calibration batch of $S$ tokens into
$\mathbf{H}\in\mathbb{R}^{S\times nd}$ with rows $\mathbf{h}_{\mathcal{B},t}^{\top}$, write
$\mathbf{W}=\mathrm{blkdiag}(\mathbf{W}_K^{\ell_1},\dots,\mathbf{W}_K^{\ell_n})\in\mathbb{R}^{nd\times np}$
for the block's true key map, so that the block's exact keys are the rows of
$\mathbf{H}\mathbf{W}$, and collect the up-projections as
$\mathbf{W}_{\uparrow}=[\mathbf{W}_{\uparrow}^{(1)},\dots,\mathbf{W}_{\uparrow}^{(n)}]\in\mathbb{R}^{R\times np}$.
The objective is
\begin{equation}
J(\mathbf{W}_{\downarrow},\mathbf{W}_{\uparrow})
=\mathbb{E}_{\mathbf{H}}\big\|\mathbf{H}\mathbf{W}
-\mathbf{H}\mathbf{W}_{\downarrow}\mathbf{W}_{\uparrow}\big\|_F^{2}.
\label{eq:objective}
\end{equation}
The data enters through one statistic, the second-moment matrix
$\mathbf{G}=\mathbb{E}[\mathbf{H}^{\top}\mathbf{H}]\in\mathbb{R}^{nd\times nd}$, whose
$(i,j)$ block is the covariance between layer $\ell_i$'s and layer $\ell_j$'s inputs.
Eq.~(\ref{eq:objective}) has a closed-form solution. Let $\mathbf{M}=\mathbf{G}^{1/2}\mathbf{W}$
with thin singular value decomposition $\mathbf{M}=\mathbf{U}\boldsymbol{\Sigma}\mathbf{V}^{\top}$.
The off-diagonal blocks of $\mathbf{G}$ are the part that BellowsKV adds. If these blocks were
zero, $\mathbf{G}$ and $\mathbf{M}$ would be block-diagonal, the singular value decomposition would
split into $n$ independent per-layer decompositions, and Eq.~(\ref{eq:objective}) would reduce to
$n$ separate per-layer low-rank caches. This per-layer construction is used by
\citet{chang2024palucompressingkvcachelowrank} and \citet{zhang2026spectra}, and it is based on the activation-aware
whitening of \citet{wang2024svdllm}. With them
present, one set of $R$ latent directions is fitted to serve all $n$ layers at once.
Then
\begin{equation}
\mathbf{W}_{\downarrow}^{\star}=\mathbf{G}^{-1/2}\,\mathbf{U}_{:R},\qquad
\mathbf{W}_{\uparrow}^{\star}=\boldsymbol{\Sigma}_{:R}\mathbf{V}_{:R}^{\top},
\label{eq:solution}
\end{equation}
attains the minimum of Eq.~(\ref{eq:objective}) with residual $\sum_{i>R}\sigma_i^{2}$
(Appendix~\ref{app:method}). The solution is unique up to an invertible mixing of
the latent coordinates, invisible at full precision but not to a quantizer; we take the square-root split
$\mathbf{W}_{\downarrow}=\mathbf{G}^{-1/2}\mathbf{U}_{:R}\boldsymbol{\Sigma}_{:R}^{1/2}$,
$\mathbf{W}_{\uparrow}=\boldsymbol{\Sigma}_{:R}^{1/2}\mathbf{V}_{:R}^{\top}$, which leaves the
latent energy-ordered and so matched to the quantizer's group-wise scales. 

\subsection{The cache at inference and the two operators}
\label{sec:decode}

At prefill, each block runs its layers on the exact keys and values, applies
Eq.~(\ref{eq:latent}) to the stacked hidden states, and stores
$(\mathbf{c}^{K}_t,\mathbf{c}^{V}_t)$ in place of the block's $2n$ per-layer tensors; a few sink
positions, a short recent window, and every token generated at decode time stay exact and
per-layer, the standard residual-window practice \citep{xiao2024streamingllm, liu2024kivi}.
Before storing, the block fits its whisper: a low-rank sketch
\citep{halko2010findingstructurerandomnessprobabilistic} of the key residual
$\mathbf{E}=\mathbf{K}-\hat{\mathbf{K}}$, one small basis per sequence and a few coefficients
per token. At decode, values never leave the latent: attention is linear in them, so each up-projection folds into the output
projection and the layer attends in $R$ dimensions
\citep{ji2025economicalinferenceenablingdeepseeks}. Keys are reconstructed through theirs,
because the positional rotation sits between latent and score
\citep{su2023roformerenhancedtransformerrotary, deepseekai2024deepseekv2strongeconomicalefficient}:
a low-rank product per layer that hides under the memory-bound cache read (settings and full
path in Appendix~\ref{app:method}).

\label{sec:operators}
BellowsKV has three parameters: the number of layers per block $n$, the number of bits per stored
value $b$, and the fraction of tokens kept $q$, and the dimension is fixed at $R=p$.

\paragraph{BellowsKV-Quant.}
We quantize the latent with uniform integer quantization, using one scale per group of $128$
adjacent coordinates. We apply no rotation before storing
\citep{ashkboos2024quarotoutlierfree4bitinference, su2025rotatekvaccuraterobust2bit} and no
per-channel treatment, because the latent is already a mixture of channels from all $n$ layers
(Observation~1). The dimension stays at $R=p$ in all settings. The main experiments use two
settings: $n=2$ with INT8 and $n=4$ with INT4, which measure $3.6\times$ and $13.8\times$ with all
metadata counted (Appendix~\ref{app:accounting}).

\paragraph{BellowsKV-Evict.}
The block stores one latent vector per token, so a token receives one eviction decision per block.
After prefill, each block scores every cached token by the attention mass that the most recent
queries place on it. The score is summed over the layers and heads of the block, and it is computed
on the keys reconstructed from the quantized latent, so that eviction uses the same keys as
decoding \citep{zhang2023h2oheavyhitteroracleefficient, li2024snapkvllmknowslooking}. The scores are then max-pooled over a window of $13$
positions, so that a token kept for its own score also keeps its neighbors \citep{li2024snapkvllmknowslooking}.
The tokens with the highest scores are kept, together with the first $4$ positions and the last
$64$ prompt positions \citep{xiao2024streamingllm}. The remaining tokens are dropped once for the
whole block, with a single gather applied to the stored latent
(Appendix~\ref{app:evict}).

\vspace{-2mm}

\section{Experiment}
\label{sec:experiments}

We evaluate BellowsKV on Llama-3.1-8B-Instruct, Mistral-7B-Instruct-v0.3, Qwen2.5-7B-Instruct, Qwen3-32B and Qwen3-30B-A3B,
with one recipe fixed throughout (Section~\ref{sec:method}): $n{=}2$ with INT8 at the low tier and
$n{=}4$ with INT4 at the high tier, no per-model tuning; the high tier is also stored at INT3 and at
K3V2 (key latent 3 bits, value latent 2) to reach and pass $16\times$. \emph{Protocol~A} reports
absolute scores on eight English LongBench tasks \citep{bai2024longbenchbilingualmultitaskbenchmark}
under the official harness; \emph{Protocol~B} reports \emph{retention}, each task's score over the
dense model's in the same harness, on 4k--32k-token inputs under the xKV authors' harness
\citep{chang2026xkvcrosslayerkvcachecompression}. \emph{Protocal~C} reports agent workload: single-turn
retrieve-then-read on BrowseComp-Plus \cite{chen2025browsecompplusfairtransparentevaluation}.

Every number is full-sample, every ratio counts
all metadata, and every table groups rows by measured ratio and marks the best per tier. Tables abbreviate BellowsKV-Quant as
\textbf{BellowsQ}, BellowsKV-Evict as \textbf{BellowsE} and the sharing-only fp16 latent as
\textbf{BellowsS}. Each subsection tests one prediction of Section~\ref{sec:motivation} where it
would fail, and the last tests the decode relation the introduction promised.

\subsection{Stream, not caches: the cross-layer head-to-head}
\label{sec:exp_crosslayer}

\begin{table}[t]
\centering
\caption{Protocol B: per-task retention of full-cache accuracy (\%) for every operating point of
every training-free cross-layer method. Rows are grouped by measured compression tier. Within each
tier, the highest compression ratio is always that of BellowsQ, so a baseline with a higher average
also stores more memory. \colorbox{blue!18}{\textbf{best}} and \colorbox{blue!7}{\underline{second}}
average per tier are marked. Task codes and settings: Appendix~\ref{app:exp_setup}.
}
\label{tab:protoB}
\vspace{3pt}
\fontsize{9pt}{9.6pt}\selectfont
\renewcommand{\arraystretch}{1.0}
\setlength{\tabcolsep}{7.4pt}
\setlength{\aboverulesep}{0.2ex}\setlength{\belowrulesep}{0.3ex}
\scalebox{0.67}{%
\begin{tabular}{@{}lrcccccccccccc@{}}
\toprule
Method & Comp. & Qsp & MFQ & HQA & MSQ & NQA & Gov & QMS & PsR & LCC & RBP & 2Wk & Avg \\
\specialrule{\lightrulewidth}{2.5pt}{3.5pt}
\multicolumn{14}{c}{\fontsize{10.5pt}{11pt}\selectfont\textbf{Llama-3.1-8B-Instruct}} \\
\addlinespace[3pt]
\midrule
MiniCache fp16 & $1.31\times$ & 72.1 & 72.3 & 74.7 & 86.8 & 55.0 & 82.7 & 89.5 & 70.7 & 80.4 & 82.6 & 55.6 & 74.8 \\
\cellcolor{blue!7}\underline{KVSharer fp16} & $1.33\times$ & 91.5 & 96.9 & 93.9 & 98.1 & 78.2 & 73.7 & 97.9 & 99.5 & 85.4 & 84.5 & 70.2 & \cellcolor{blue!7}\underline{88.2} \\
\cellcolor{blue!18}\textbf{BellowsQ $n{=}2$ INT8} & $3.64\times$ & 95.6 & 97.6 & 99.6 & 99.4 & 95.9 & 98.5 & 102.5 & 100.0 & 98.7 & 99.0 & 101.0 & \cellcolor{blue!18}\textbf{98.9} \\
\midrule
CommonKV & $3.95\times$ & 92.2 & 90.6 & 98.0 & 100.1 & 89.0 & 89.7 & 96.5 & 99.5 & 98.4 & 97.8 & 91.1 & 94.8 \\
xKV & $4.00\times$ & 96.6 & 98.6 & 96.0 & 102.1 & 97.9 & 97.0 & 100.8 & 100.0 & 99.4 & 99.4 & 92.1 & 98.2 \\
MiniCache+INT4 & $4.78\times$ & 70.2 & 70.9 & 70.4 & 80.2 & 51.8 & 81.1 & 89.9 & 70.8 & 79.8 & 82.7 & 56.4 & 73.1 \\
KVSharer+INT4 & $4.99\times$ & 91.1 & 94.8 & 94.0 & 99.5 & 79.7 & 72.4 & 97.8 & 99.5 & 85.2 & 83.3 & 70.0 & 87.9 \\
XQuant published & $6.59\times$ & 46.8 & 64.5 & 71.0 & 59.1 & 49.6 & 69.3 & 82.0 & 6.0 & 57.2 & 64.5 & 42.4 & 55.7 \\
MiniCache all+INT4 & $6.72\times$ & 13.4 & 13.2 & 9.6 & 13.9 & 7.5 & 20.7 & 50.6 & 30.1 & 51.4 & 58.3 & 4.4 & 24.8 \\
\cellcolor{blue!18}\textbf{BellowsQ $n{=}2$ INT4} & $6.90\times$ & 96.3 & 97.7 & 100.3 & 97.6 & 98.6 & 98.6 & 101.1 & 100.0 & 99.6 & 99.2 & 101.6 & \cellcolor{blue!18}\textbf{99.2} \\
CommonKV & $7.89\times$ & 86.0 & 82.6 & 91.1 & 92.7 & 85.4 & 77.8 & 90.9 & 99.2 & 81.2 & 88.5 & 78.1 & 86.7 \\
\cellcolor{blue!7}\underline{BellowsQ $n{=}3$ INT4} & $10.40\times$ & 93.8 & 96.7 & 99.1 & 102.5 & 97.2 & 97.0 & 99.9 & 100.0 & 99.4 & 98.9 & 100.3 & \cellcolor{blue!7}\underline{98.6} \\
\midrule
\cellcolor{blue!7}\underline{KVSharer+INT2} & $11.24\times$ & 53.6 & 67.4 & 63.9 & 79.0 & 69.2 & 30.8 & 78.2 & 57.8 & 59.7 & 59.3 & 37.0 & \cellcolor{blue!7}\underline{59.6} \\
MiniCache+INT2 & $11.96\times$ & 11.1 & 13.1 & 5.4 & 10.4 & 9.4 & 20.0 & 39.6 & 11.3 & 46.4 & 56.0 & 4.0 & 20.6 \\
\cellcolor{blue!18}\textbf{BellowsQ $n{=}4$ INT4} & $13.84\times$ & 94.6 & 99.0 & 99.1 & 96.4 & 96.2 & 93.5 & 98.7 & 100.0 & 99.7 & 99.0 & 101.5 & \cellcolor{blue!18}\textbf{98.0} \\
\midrule
XQuant 1-bit & $15.11\times$ & 0.0 & 0.1 & 0.0 & 0.0 & 0.0 & 0.0 & 0.0 & 0.0 & 12.3 & 15.4 & 0.0 & 2.5 \\
CommonKV & $15.78\times$ & 34.6 & 49.7 & 43.0 & 55.6 & 56.5 & 50.5 & 83.0 & 77.9 & 54.4 & 56.8 & 28.0 & 53.6 \\
xKV & $16.00\times$ & 91.4 & 80.9 & 82.4 & 87.3 & 73.3 & 75.6 & 94.4 & 95.5 & 92.8 & 86.8 & 58.5 & 83.5 \\
\cellcolor{blue!18}\textbf{BellowsQ $n{=}4$ INT3} & $17.65\times$ & 94.2 & 98.6 & 98.0 & 101.3 & 98.8 & 93.6 & 98.0 & 100.0 & 101.1 & 98.9 & 99.5 & \cellcolor{blue!18}\textbf{98.4} \\
\cellcolor{blue!7}\underline{BellowsQ $n{=}4$ K3V2} & $20.47\times$ & 91.6 & 96.6 & 97.9 & 100.4 & 99.6 & 93.5 & 98.0 & 100.0 & 99.9 & 99.3 & 96.6 & \cellcolor{blue!7}\underline{97.6} \\
\specialrule{\heavyrulewidth}{5pt}{1.4pt}
\specialrule{\lightrulewidth}{0pt}{3.5pt}
\multicolumn{14}{c}{\fontsize{10.5pt}{11pt}\selectfont\textbf{Mistral-7B-Instruct-v0.3}} \\
\addlinespace[3pt]
\midrule
MiniCache fp16 & $1.31\times$ & 76.7 & 69.4 & 72.7 & 56.2 & 81.4 & 68.7 & 83.3 & 37.2 & 76.3 & 74.6 & 67.4 & 69.4 \\
\cellcolor{blue!7}\underline{KVSharer fp16} & $1.33\times$ & 103.8 & 95.6 & 94.2 & 90.6 & 100.3 & 91.8 & 93.3 & 68.4 & 89.1 & 87.7 & 87.4 & \cellcolor{blue!7}\underline{91.1} \\
\cellcolor{blue!18}\textbf{BellowsQ $n{=}2$ INT8} & $3.64\times$ & 98.2 & 100.6 & 98.9 & 102.1 & 95.9 & 99.4 & 99.7 & 99.5 & 100.4 & 101.0 & 100.3 & \cellcolor{blue!18}\textbf{99.6} \\
\midrule
CommonKV & $3.95\times$ & 99.4 & 101.2 & 96.4 & 89.7 & 92.3 & 89.7 & 95.8 & 100.0 & 97.1 & 100.0 & 100.0 & 96.5 \\
\cellcolor{blue!18}\textbf{xKV} & $4.00\times$ & 102.0 & 100.0 & 100.2 & 100.4 & 95.2 & 98.2 & 99.5 & 95.9 & 98.3 & 100.2 & 103.2 & \cellcolor{blue!18}\textbf{99.4} \\
MiniCache+INT4 & $4.78\times$ & 79.0 & 68.6 & 70.6 & 52.7 & 81.4 & 67.9 & 84.0 & 38.8 & 76.5 & 74.8 & 65.4 & 69.1 \\
KVSharer+INT4 & $4.99\times$ & 101.1 & 97.1 & 95.1 & 90.2 & 92.9 & 92.5 & 93.2 & 67.9 & 87.6 & 88.0 & 89.0 & 90.4 \\
XQuant published & $6.59\times$ & 59.1 & 67.7 & 72.8 & 51.1 & 69.9 & 74.9 & 86.9 & 13.3 & 68.2 & 70.4 & 73.5 & 64.3 \\
MiniCache all+INT4 & $6.72\times$ & 35.2 & 23.3 & 22.1 & 23.4 & 39.1 & 18.9 & 50.7 & 19.0 & 49.5 & 42.8 & 30.4 & 32.2 \\
\cellcolor{blue!7}\underline{BellowsQ $n{=}2$ INT4} & $7.02\times$ & 98.4 & 98.8 & 98.3 & 101.5 & 95.6 & 98.3 & 99.4 & 98.0 & 99.8 & 100.8 & 101.6 & \cellcolor{blue!7}\underline{99.1} \\
CommonKV & $7.90\times$ & 90.9 & 93.4 & 98.1 & 84.1 & 101.5 & 79.1 & 95.7 & 95.9 & 93.7 & 93.4 & 94.6 & 92.8 \\
BellowsQ $n{=}3$ INT4 & $10.46\times$ & 96.6 & 100.8 & 99.7 & 100.5 & 92.1 & 94.9 & 96.7 & 96.9 & 98.5 & 99.9 & 99.5 & 97.8 \\
\midrule
\cellcolor{blue!7}\underline{KVSharer 37.5\%+INT2} & $11.24\times$ & 90.3 & 79.8 & 82.4 & 73.5 & 89.9 & 78.3 & 88.1 & 17.3 & 77.3 & 69.4 & 70.1 & \cellcolor{blue!7}\underline{74.2} \\
MiniCache+INT2 & $11.99\times$ & 34.6 & 18.8 & 21.2 & 23.3 & 28.8 & 21.4 & 45.4 & 11.2 & 43.2 & 40.9 & 23.0 & 28.3 \\
\cellcolor{blue!18}\textbf{BellowsQ $n{=}4$ INT4} & $13.84\times$ & 93.9 & 103.1 & 99.1 & 97.2 & 96.3 & 93.6 & 99.8 & 97.4 & 98.6 & 101.0 & 99.3 & \cellcolor{blue!18}\textbf{98.1} \\
\midrule
KVSharer+INT2 & $14.05\times$ & 59.3 & 67.9 & 74.6 & 56.6 & 66.2 & 46.5 & 74.6 & 13.9 & 54.9 & 43.2 & 61.7 & 56.3 \\
XQuant 1-bit & $15.11\times$ & 0.0 & 0.5 & 0.0 & 0.0 & 0.0 & 0.1 & 0.0 & 0.0 & 17.2 & 13.0 & 0.0 & 2.8 \\
CommonKV & $15.80\times$ & 53.1 & 55.8 & 24.9 & 21.1 & 48.8 & 60.2 & 81.0 & 68.9 & 66.2 & 64.3 & 32.0 & 52.4 \\
xKV & $16.00\times$ & 90.4 & 88.5 & 47.8 & 42.1 & 85.2 & 93.0 & 91.8 & 81.7 & 92.7 & 90.0 & 71.1 & 79.5 \\
\cellcolor{blue!7}\underline{BellowsQ $n{=}4$ INT3} & $17.65\times$ & 94.8 & 101.0 & 99.7 & 95.9 & 96.0 & 93.8 & 97.7 & 96.4 & 97.9 & 101.7 & 100.5 & \cellcolor{blue!7}\underline{97.8} \\
\cellcolor{blue!18}\textbf{BellowsQ $n{=}4$ K3V2} & $20.47\times$ & 96.8 & 100.5 & 99.6 & 95.0 & 97.0 & 93.7 & 99.5 & 94.4 & 99.0 & 101.9 & 99.1 & \cellcolor{blue!18}\textbf{97.9} \\
\specialrule{\heavyrulewidth}{5pt}{1.4pt}
\specialrule{\lightrulewidth}{0pt}{3.5pt}
\multicolumn{14}{c}{\fontsize{10.5pt}{11pt}\selectfont\textbf{Qwen2.5-7B-Instruct}} \\
\addlinespace[3pt]
\midrule
MiniCache fp16 & $1.31\times$ & 25.2 & 28.8 & 19.9 & 18.9 & 25.1 & 55.7 & 66.1 & 3.2 & 42.5 & 44.2 & 15.8 & 31.4 \\
\cellcolor{blue!7}\underline{KVSharer fp16} & $1.33\times$ & 51.5 & 61.5 & 36.5 & 31.5 & 25.4 & 38.9 & 68.3 & 82.8 & 67.0 & 57.5 & 26.1 & \cellcolor{blue!7}\underline{49.7} \\
\cellcolor{blue!18}\textbf{BellowsQ $n{=}2$ INT8} & $3.53\times$ & 101.0 & 99.1 & 100.4 & 98.1 & 97.0 & 97.9 & 99.8 & 100.0 & 96.6 & 97.1 & 100.0 & \cellcolor{blue!18}\textbf{98.8} \\
\midrule
CommonKV & $3.95\times$ & 93.0 & 89.6 & 94.6 & 86.9 & 99.9 & 86.2 & 92.6 & 98.5 & 89.8 & 95.0 & 88.9 & 92.3 \\
\cellcolor{blue!18}\textbf{xKV$^\dagger$} & $4.00\times$ & 97.9 & 101.9 & 96.9 & -- & 109.8 & 97.2 & 102.1 & 100.0 & 92.3 & 96.1 & 96.8 & \cellcolor{blue!18}\textbf{99.1} \\
MiniCache+INT4 & $4.78\times$ & 24.8 & 30.1 & 20.4 & 17.2 & 24.8 & 56.5 & 64.3 & 4.0 & 44.2 & 43.3 & 13.8 & 31.2 \\
KVSharer+INT4 & $4.99\times$ & 60.2 & 60.7 & 36.1 & 33.3 & 30.4 & 41.4 & 67.5 & 82.1 & 63.5 & 56.3 & 22.9 & 50.4 \\
XQuant published & $6.59\times$ & 34.0 & 39.8 & 41.4 & 27.5 & 29.3 & 50.0 & 75.1 & 4.1 & 30.9 & 35.4 & 28.8 & 36.0 \\
\cellcolor{blue!7}\underline{BellowsQ $n{=}2$ INT4} & $6.66\times$ & 101.6 & 100.2 & 99.2 & 97.3 & 97.2 & 97.3 & 100.5 & 100.0 & 97.7 & 98.1 & 98.6 & \cellcolor{blue!7}\underline{98.9} \\
MiniCache all+INT4 & $6.72\times$ & 14.1 & 14.4 & 10.8 & 8.7 & 17.3 & 34.6 & 63.0 & 0.2 & 36.2 & 35.5 & 9.8 & 22.2 \\
CommonKV & $7.89\times$ & 59.0 & 59.0 & 60.8 & 49.9 & 73.7 & 66.3 & 88.6 & 72.5 & 69.5 & 70.5 & 35.0 & 64.1 \\
xKV$^\dagger$ & $8.00\times$ & 94.3 & 96.6 & 85.7 & 78.9 & 73.8 & 93.7 & 99.4 & 98.5 & 91.9 & -- & 85.0 & 89.8 \\
BellowsQ $n{=}3$ INT4 & $9.92\times$ & 100.0 & 96.5 & 99.5 & 98.4 & 101.6 & 94.3 & 100.5 & 99.5 & 95.3 & 96.1 & 100.7 & 98.4 \\
\midrule
\cellcolor{blue!7}\underline{KVSharer+INT2} & $10.93\times$ & 19.0 & 30.9 & 25.3 & 16.6 & 25.9 & 32.0 & 50.8 & 15.0 & 36.0 & 36.9 & 9.9 & \cellcolor{blue!7}\underline{27.1} \\
MiniCache+INT2 & $11.97\times$ & 13.0 & 12.6 & 8.3 & 7.6 & 17.9 & 39.9 & 56.9 & 1.4 & 35.3 & 30.9 & 7.4 & 21.0 \\
\cellcolor{blue!18}\textbf{BellowsQ $n{=}4$ INT4} & $13.13\times$ & 100.8 & 97.3 & 98.6 & 99.2 & 96.8 & 93.6 & 99.9 & 99.0 & 92.6 & 94.5 & 98.6 & \cellcolor{blue!18}\textbf{97.3} \\
\midrule
XQuant 1-bit & $15.11\times$ & 0.7 & 2.1 & 0.1 & 0.6 & 1.3 & 9.8 & 12.7 & 1.0 & 21.2 & 24.1 & 0.1 & 6.7 \\
CommonKV & $15.78\times$ & 15.2 & 22.9 & 31.1 & 22.6 & 22.9 & 47.7 & 66.4 & 5.5 & 38.6 & 42.8 & 11.7 & 29.8 \\
xKV & $16.00\times$ & 63.0 & 69.6 & 26.1 & 22.2 & 38.2 & 80.8 & 78.3 & 48.5 & 74.7 & 61.5 & 35.9 & 54.4 \\
\cellcolor{blue!18}\textbf{BellowsQ $n{=}4$ INT3} & $16.52\times$ & 98.2 & 98.8 & 98.6 & 96.7 & 97.2 & 93.3 & 101.4 & 98.0 & 94.5 & 93.3 & 97.1 & \cellcolor{blue!18}\textbf{97.0} \\
\cellcolor{blue!7}\underline{BellowsQ $n{=}4$ K3V2} & $18.96\times$ & 99.6 & 95.8 & 99.6 & 98.8 & 100.0 & 93.2 & 100.0 & 98.0 & 88.4 & 94.6 & 96.3 & \cellcolor{blue!7}\underline{96.8} \\
\bottomrule
\end{tabular}
}
\vspace{3pt}
\end{table}

Section~\ref{sec:ob1} predicts that reusing one layer's raw keys and values in another fails and
storing a latent of the stream does not, at any ratio. Protocol~B runs every training-free
cross-layer method we know of, placed by measured bytes
(Table~\ref{tab:protoB}, Appendix Figures~\ref{fig:pareto_llama}--\ref{fig:pareto_qwen}): xKV
\citep{chang2026xkvcrosslayerkvcachecompression} and CommonKV
\citep{wang2025commonkvcompressingkvcache} store a latent; MiniCache
\citep{liu2024minicachekvcachecompression}, KVSharer \citep{yang2024kvsharer} and XQuant
\citep{yang2025xquantcrosslayer} share raw caches. The raw-cache routes are broken at $1.3\times$,
before any bits are cut; every latent method works at $\sim$4$\times$; at $\sim$16$\times$ xKV
retains at most 84\% and CommonKV 54\%, while BellowsQ retains 97--98\% at $13$--$14\times$, at
INT3 (fewer bytes than xKV's point) and at $19$--$20.5\times$, where no cross-layer point
exists. The raw-cache methods already lose accuracy at $1.3\times$, which shows that the caches themselves are not the object to share. Our points do not fall with xKV's at high compression, which shows that sharing itself is not the cause of the loss. 

\paragraph{Beyond LongBench.}
Protocol~C repeats the comparison on an agent task that must read its retrieved
evidence verbatim, on two models the paper has not used, one of them a mixture of
experts (Table~\ref{tab:protoc}), and there the rank cliff arrives two octaves
earlier: xKV scores $26.5$ at $8\times$ and, on the mixture of experts, $15.5$ at
$4\times$, and CommonKV has lost half of dense by $8\times$. Cutting bits at full rank
still costs nothing, with BellowsQ at $86.0$ and $85.8$ at $14.6\times$ against dense
$85.5$ and $86.1$, so the rule holds where the task is least forgiving.

\begin{table}[t]
\centering
\scriptsize
\caption{Protocol~C: single-turn retrieve-then-read on BrowseComp-Plus on all 830 queries, with the oracle prompt at a 32{,}768-token budget. Results are scored by the benchmark's own judge. The notation ``--'' means that the method cannot reach that compression rate. See full table in Appendix~\ref{app:protoC_full}}
\label{tab:protoc}
\vspace{2pt}
\setlength{\tabcolsep}{4.5pt}
\begin{tabular}{@{}llccccccc@{}}
\toprule
& & & \multicolumn{3}{c}{cross-layer latent} & \multicolumn{3}{c}{per-layer quantized} \\
\cmidrule(lr){4-6}\cmidrule(lr){7-9}
Model & Tier & Dense & \textbf{BellowsQ} & CommonKV & xKV & OTT & KIVI & PolarQuant \\
\midrule
\multirow{2}{*}{Qwen3-32B}
 & $\sim$8$\times$  & 85.54 & \textbf{84.22} & 81.08 & 26.51 & 77.95 & 75.42 & 51.93 \\
 & $\sim$16$\times$ & 85.54 & \textbf{86.02} & 66.02 & \phantom{0}8.43 & -- & -- & -- \\
\midrule
\multirow{2}{*}{Qwen3-30B-A3B}
 & $\sim$8$\times$  & 86.14 & \textbf{85.30} & 49.88 & \phantom{0}5.30 & 81.45 & 77.59 & \phantom{0}4.22 \\
 & $\sim$16$\times$ & 86.14 & \textbf{85.78} & -- & -- & -- & -- & -- \\
\bottomrule
\end{tabular}
\end{table}

\vspace{-2mm}
\subsection{Dimension and bit-width at a fixed budget}
\label{sec:exp_rank}

\begin{table}[t]
\centering
\caption{Cutting rank is the cliff; the rival that holds its rank and cuts bits instead comes back. (See full table in Appendix~\ref{app:tab:transplant})}
\label{tab:rank}
\vspace{3pt}
\scriptsize
\renewcommand{\arraystretch}{1.0}
\setlength{\tabcolsep}{4pt}
\begin{tabular}{@{}lcccc@{}}
\toprule
Arm & Comp. & Llama & Mistral & Qwen \\
\midrule
xKV, fp16, rank $p$ & $4.0\times$ & 98.2 & 99.4 & 99.1 \\
xKV, fp16, rank $p/2$ & $8.0\times$ & -- & -- & 89.8 \\
xKV, fp16, rank $p/4$ & $16.0\times$ & 83.5 & 79.5 & 54.4 \\
\midrule
\textbf{xKV-hold}: the same basis, near-full rank, our INT4 leg & $15.1\times$ & 97.6 & 99.2 & 23.4 \\
\midrule
BellowsQ $n{=}4$ INT4 (the frozen consensus) & $13.1$--$13.8\times$ & 98.0 & 98.1 & 97.3 \\
our construction, per-sequence SVD oracle ($2.2\times$ prefill) & $15.1\times$ & 97.3 & -- & -- \\
\midrule
CommonKV, rank $81\%$ of the width & $4.0\times$ & 94.8 & 96.5 & 92.3 \\
CommonKV, rank $20\%$ of the width & $15.8\times$ & 53.6 & 52.4 & 29.8 \\
\bottomrule
\end{tabular}
\vspace{2pt}
\end{table}

We compare two ways to reach a high compression ratio at a fixed memory budget: reducing the
dimension of the shared basis, and reducing the number of bits per stored value. We run this
comparison on the construction of xKV, because both settings exist for it (Table~\ref{tab:rank}).
xKV reaches $16\times$ by reducing its per-sequence basis from rank $p$ to $p/4$. xKV-hold keeps
the same basis at near-full rank and stores it at INT4, which gives $15.1\times$. A third setting
replaces our fixed projection with a per-sequence SVD, at $2.2\times$ the prefill cost.

xKV-hold reaches $97.6$ on Llama and $99.2$ on Mistral, against $83.5$ and $79.5$ for xKV at
$16\times$. The per-sequence SVD reaches $97.3$ on Llama, which is below the $98.0$ of the fixed
projection. On Qwen, every setting that stores xKV's basis at four bits fails, and we have no
explanation for this case. The results indicate that the loss of xKV at $16\times$ is caused by the
reduced dimension. xKV-hold keeps the same shared basis, so the sharing itself is not the cause.

\vspace{-2mm}

\subsection{Comparison with per-layer quantizers}
\label{sec:exp_protoA}

\begin{table}[t]
\centering
\caption{Protocol A: LongBench average over eight tasks under the official harness. The table
reports every official configuration of KIVI, OTT, RotateKV, TurboQuant and PolarQuant, and every
BellowsKV point, grouped by measured compression tier. Within every tier, the highest compression ratio is that of BellowsKV,
so a per-layer quantizer with a higher average also stores more memory. No per-layer point exists
above $8\times$, because two bits at full dimension already give $8\times$. Per-task results:
Appendix~\ref{app:protoA_full}. Equal bit-width comparison: Table~\ref{app:tab:eqbw}.}
\label{tab:protoA}
\vspace{3pt}
\fontsize{6.8pt}{7.6pt}\selectfont
\renewcommand{\arraystretch}{1.0}
\setlength{\tabcolsep}{1.8pt}
\makeatletter\@ifundefined{pameth}{\newlength{\pameth}\newlength{\pacomp}}{}\makeatother
\settowidth{\pameth}{BellowsQ $n{=}2$ INT8\,}
\settowidth{\pacomp}{$12.2$--$13.0\times$\,}
\begin{tabular}[t]{@{}lrccc@{}}
\toprule
\makebox[\pameth][l]{Method} & \makebox[\pacomp][r]{Comp.} & Llama & Mistral & Qwen \\
\midrule
\makebox[\pameth][l]{Dense fp16} & \makebox[\pacomp][r]{$1.00\times$} & 53.29 & 52.78 & 53.23 \\
\midrule
\makebox[\pameth][l]{BellowsS $n{=}2$} & \makebox[\pacomp][r]{$1.9\times$} & 53.07 & \cellcolor{blue!18}\textbf{53.04} & 51.88 \\
\makebox[\pameth][l]{BellowsS $n{=}3$} & \makebox[\pacomp][r]{$2.9$--$2.9\times$} & 52.67 & 52.87 & 51.92 \\
\makebox[\pameth][l]{BellowsQ $n{=}2$ INT8} & \makebox[\pacomp][r]{$3.6$--$3.7\times$} & 53.05 & 52.92 & \cellcolor{blue!7}\underline{52.00} \\
\makebox[\pameth][l]{PolarQuant r4t4v4} & \makebox[\pacomp][r]{$3.7\times$} & 53.66 & -- & \cellcolor{blue!18}\textbf{53.90} \\
\makebox[\pameth][l]{KIVI K4V4} & \makebox[\pacomp][r]{$3.8\times$} & \cellcolor{blue!18}\textbf{53.92} & 52.80 & -- \\
\makebox[\pameth][l]{OTT K4V4} & \makebox[\pacomp][r]{$3.8\times$} & \cellcolor{blue!7}\underline{53.83} & 52.99 & -- \\
\makebox[\pameth][l]{BellowsS $n{=}4$} & \makebox[\pacomp][r]{$3.8$--$3.9\times$} & 52.78 & \cellcolor{blue!7}\underline{53.04} & 51.21 \\
\midrule
\makebox[\pameth][l]{TurboQuant K4V4} & \makebox[\pacomp][r]{$4.0\times$} & \cellcolor{blue!7}\underline{52.84} & 52.44 & \cellcolor{blue!7}\underline{52.98} \\
\makebox[\pameth][l]{RotateKV 4-bit} & \makebox[\pacomp][r]{$4.0\times$} & -- & \cellcolor{blue!18}\textbf{53.14} & -- \\
\makebox[\pameth][l]{PolarQuant r3t3v4} & \makebox[\pacomp][r]{$4.1\times$} & \cellcolor{blue!18}\textbf{53.50} & -- & \cellcolor{blue!18}\textbf{53.89} \\
\makebox[\pameth][l]{PolarQuant r2t2v4} & \makebox[\pacomp][r]{$4.7\times$} & 50.63 & -- & 50.87 \\
\makebox[\pameth][l]{RotateKV 3-bit} & \makebox[\pacomp][r]{$5.3\times$} & 52.38 & 52.33 & -- \\
\makebox[\pameth][l]{TurboQuant K4V2} & \makebox[\pacomp][r]{$5.3\times$} & 52.57 & 51.84 & 51.25 \\
\makebox[\pameth][l]{BellowsQ $n{=}3$ INT8} & \makebox[\pacomp][r]{$5.3$--$5.5\times$} & 52.84 & \cellcolor{blue!7}\underline{53.04} & 51.90 \\
\bottomrule
\end{tabular}
\hspace{12pt}
\begin{tabular}[t]{@{}lrccc@{}}
\toprule
\makebox[\pameth][l]{Method} & \makebox[\pacomp][r]{Comp.} & Llama & Mistral & Qwen \\
\midrule
\makebox[\pameth][l]{BellowsQ $n{=}2$ INT4} & \makebox[\pacomp][r]{$6.4$--$6.8\times$} & \cellcolor{blue!7}\underline{52.92} & \cellcolor{blue!18}\textbf{52.76} & \cellcolor{blue!18}\textbf{51.86} \\
\makebox[\pameth][l]{PolarQuant r2t2v2} & \makebox[\pacomp][r]{$6.7\times$} & 48.51 & -- & 50.12 \\
\makebox[\pameth][l]{KIVI K2V2} & \makebox[\pacomp][r]{$7.1\times$} & 52.17 & 52.13 & -- \\
\makebox[\pameth][l]{OTT K2V2} & \makebox[\pacomp][r]{$7.1\times$} & \cellcolor{blue!18}\textbf{53.30} & 52.51 & -- \\
\makebox[\pameth][l]{TurboQuant K2V2} & \makebox[\pacomp][r]{$8.0\times$} & 46.43 & 48.31 & 47.18 \\
\makebox[\pameth][l]{RotateKV 2-bit} & \makebox[\pacomp][r]{$8.0\times$} & 47.00 & 49.01 & -- \\
\makebox[\pameth][l]{BellowsQ $n{=}2$ INT3} & \makebox[\pacomp][r]{$8.0$--$8.6\times$} & 52.72 & \cellcolor{blue!7}\underline{52.67} & \cellcolor{blue!7}\underline{51.60} \\
\midrule
\makebox[\pameth][l]{BellowsQ $n{=}4$ INT4} & \makebox[\pacomp][r]{$12.2$--$13.0\times$} & \cellcolor{blue!18}\textbf{52.45} & \cellcolor{blue!7}\underline{52.83} & \cellcolor{blue!18}\textbf{51.12} \\
\makebox[\pameth][l]{BellowsQ $n{=}4$ INT3} & \makebox[\pacomp][r]{$15.0$--$16.2\times$} & \cellcolor{blue!7}\underline{52.42} & 52.75 & 50.02 \\
\makebox[\pameth][l]{BellowsQ $n{=}4$ K3V2} & \makebox[\pacomp][r]{$16.9$--$18.6\times$} & 52.27 & \cellcolor{blue!18}\textbf{52.87} & \cellcolor{blue!7}\underline{50.28} \\
\bottomrule
\end{tabular}
\vspace{2pt}
\end{table}

Protocol~A compares the shared latent with per-layer quantizers at matched memory. We run every
official configuration of KIVI, OTT, RotateKV, TurboQuant and PolarQuant
\citep{liu2024kivi, su2025ottaccuratekvcache, su2025rotatekvaccuraterobust2bit,
zandieh2025turboquantonlinevectorquantization, han2025polarquantquantizingkvcaches}, and every
BellowsKV point from $1.92\times$ to $18.6\times$ (Table~\ref{tab:protoA}).

On the latent at $n{=}4$, the loss from fp16 to INT4 is $0.1$ to $0.3$ points, and the loss from
INT4 to INT3 is at most $1.1$ points. For the rotated per-layer caches, the loss from four bits to
two bits is $4.1$ to $6.4$ points. Note that these two comparisons do not use the same change in
bit-width; the comparison at equal bit-width is given in Table~\ref{app:tab:eqbw}. At
$\sim$7$\times$ on Llama, OTT at two bits reaches $53.30$ against $52.92$ for BellowsKV, while
storing $5\%$ fewer bytes. Above $8\times$, no per-layer configuration exists, because two bits at
full dimension already gives $8\times$.

\label{sec:exp_evict}

\begin{table}[t]
\centering
\caption{Protocol B retention (\%) of full-cache accuracy. All routes use the same scoring rule.
The per-layer baseline applies this rule independently in each layer, on the fp16 cache of that
layer. The comparison at equal keep ratio is given in Table~\ref{app:tab:evict_tiers}.}
\label{tab:evict}
\vspace{3pt}
\scriptsize
\renewcommand{\arraystretch}{1.0}
\begin{tabular}{@{}lllcccc@{}}
\toprule
Budget & Route & Keep & Comp. & Llama & Mistral & Qwen \\
\midrule
\multirow{2}{*}{$\sim$8$\times$} & BellowsE & 1/2 & 7.7--7.8 & \textbf{97.6} & \textbf{97.9} & \textbf{99.2} \\
 & Per-layer eviction & 1/8 & 8.0 & 94.7 & 94.9 & 98.3 \\
\midrule
\multirow{2}{*}{$\sim$16$\times$} & BellowsE & 1/4 & 15.2--15.4 & \textbf{96.1} & \textbf{96.4} & \textbf{97.9} \\
 & Per-layer eviction & 1/16 & 16.0 & 89.5 & 90.7 & 92.6 \\
\midrule
\multirow{3}{*}{$\sim$30$\times$} & BellowsQ$\times$BellowsE & 1/2 & 26.3--27.7 & \textbf{97.5} & 97.9 & \textbf{99.2} \\
 & BellowsE & 1/8 & 29.6--30.0 & 92.5 & 93.8 & 95.0 \\
 & xKV$+$INT4$+$pooled SnapKV & 1/2 & 30.1 & 96.1 & \textbf{98.4} & 23.8 \\
\bottomrule
\end{tabular}
\vspace{3pt}
\end{table}

We also compare one eviction decision per block with the same rule applied independently in each layer
(Table~\ref{tab:evict}). Both routes use the same scoring rule. The per-layer baseline scores each
layer on its own fp16 cache. The full grid is given in Appendix~\ref{app:evict_full}.

At the same keep ratio, the per-layer baseline is better by $0.8$ to $3.3$ points, and it stores
$3.8\times$ more memory. At equal memory, the block-level decision is better by $0.9$ to $3.0$
points at about $8\times$, and by $5.3$ to $6.6$ points at about $16\times$. The advantage is
larger at the higher compression ratio.

\vspace{-2mm}
\subsection{Decode cost}
\label{sec:exp_decode}

The introduction promised a decode cost, with a relation bounding any reconstruct-at-read
design: decode speedup $=(d_{kv}/R)\times 2/(\text{bytes per latent value at read})$. $n$ is
absent, so sharing buys capacity, never speed, and a design holding $R{=}d_{kv}$ that reads fp16
should decode at parity. On an H100 (HuggingFace, 32k context, no custom kernel) BellowsKV
decodes at $0.995\times$ dense against a predicted $1.00\times$; Palu
\citep{chang2024palucompressingkvcachelowrank}, the rank-cutting design with a published decode
figure, reports $1.89\times$ against a predicted $2.00\times$ (Appendix Table~\ref{app:tab:decode}).
The freed bytes buy a $2\times$ batch under HuggingFace and a $14.8\times$ KV pool under vLLM
(above the $13.8\times$ byte ratio for a reason we have not isolated). The measured decode speed is $0.995\times$ that of the uncompressed cache. This confirms two
properties of the design: reconstruction at read adds no measurable cost, and sharing across layers
does not change decode speed.

\vspace{-2mm}

\section{Conclusion}
\label{sec:conclusion}
\vspace{-2mm}

BellowsKV starts from one measured property: what a block of layers shares is the stream
behind its caches, and the aggregate drawn from it is a consensus, tame in amplitude, reliable
in judgment, irreducible in dimension. One training-free construction turns it into a
compressor whose two operators compose: near-dense accuracy at ratios where published
cross-layer methods fail, at the decode speed of an uncompressed cache. The takeaway is the
rule the experiments price: hold the rank, buy bits first, evict last.

\clearpage

\bibliography{iclr2027_conference}
\bibliographystyle{iclr2027_conference}

\newpage
\appendix
\section{Related works}
\label{app:related}

Work on the key-value cache divides by the object it acts on. One line keeps every entry and stores
each number in fewer bits. A second line keeps full precision and stores fewer tokens. A third line
changes what is stored at all and asks several layers to share one object. BellowsKV belongs to the
third line, and this appendix places it against all three.

\subsection{Quantizing the cache}
\label{app:rel_quant}

The most direct way to shrink a cache is to keep every entry and spend fewer bits on each number.
What makes this hard is that the cache carries channel-aligned outliers, which are a few coordinates
whose dynamic range forces a shared scale that crushes everything else. One group of methods protects
those coordinates. KIVI \citep{liu2024kivi} quantizes keys along the channel axis and values along the
token axis, so that the numbers sharing a scale are comparable to begin with, and it reaches two bits
without any training. KVQuant \citep{hooper2024kvquant} calibrates a non-uniform grid per channel and
keeps a sparse set of outliers in full precision, which is what allows it to report context lengths in
the millions. OTT \citep{su2025ottaccuratekvcache} tracks which tokens carry the outliers as decoding
proceeds and stores those exactly, and it is the one baseline in our Protocol-A field that still works
at two bits.

A second group removes the outliers instead of protecting them. QuaRot
\citep{ashkboos2024quarotoutlierfree4bitinference} multiplies the cache by an orthogonal matrix before
quantizing, so that every stored coordinate becomes a mixture of all the others and no single channel
can dominate a scale. SpinQuant \citep{liu2025spinquantllmquantizationlearned} learns that rotation
rather than drawing it at random, and RotateKV \citep{su2025rotatekvaccuraterobust2bit} adapts it to the
outlier structure of each head. PolarQuant \citep{han2025polarquantquantizingkvcaches} changes the
coordinate system instead of rotating inside it and codes each key by its direction and its magnitude,
while TurboQuant \citep{zandieh2025turboquantonlinevectorquantization} replaces scalar rounding with
online vector quantization at a near-optimal distortion rate. The most recent work treats the whole
problem as transform coding and fits the transform to the model's own activation statistics offline,
which is the shared idea behind OSCAR \citep{zhou2026oscarofflinespectralcovarianceaware}, OScaR
\citep{su2026oscaroccamsrazorextreme} and SPECTRA \citep{zhang2026spectra}, and
\citet{staniszewski2026kvcachetransformcoding} state the coding view explicitly. Every method in this
paragraph acts inside a single layer, and every one of them applies its transform again at each read.
The latent BellowsKV stores is already a dense mixture of the channels of several layers, so it needs no
rotation before storage, and Section~\ref{sec:exp_protoA} tests that claim against each of these methods
whose released code we could run.

\subsection{Evicting tokens}
\label{app:rel_evict}

The second line keeps the stored numbers exact and drops tokens instead. H2O
\citep{zhang2023h2oheavyhitteroracleefficient} observes that a small set of heavy hitters receives most
of the attention mass over a long prefix and keeps those tokens. SnapKV
\citep{li2024snapkvllmknowslooking} scores each cached token by the attention that the most recent
queries place on it and then pools the profile over a short window, so that a token kept for its own
score also keeps its neighbours. StreamingLLM \citep{xiao2024streamingllm} shows that the first few
positions act as attention sinks and must never be dropped, which is why nearly every later method,
including ours, exempts a sink region and a recent window. Eviction is the one axis whose damage cannot
be undone at read time, because a token that has been dropped is gone for every later query.

All of these methods decide layer by layer, since each layer owns its cache and is free to rank tokens
as it likes. A smaller body of work asks whether that freedom is worth its cost.
\citet{ma2025compressingkvcachelongcontext} report that neighbouring layers attend to much the same
tokens and reuse one layer's decision in the next, and HShare \citep{wu2025hshare} organises the same
redundancy hierarchically across layers, heads and queries in order to speed up decoding. Our eviction
operator takes this reading to its conclusion. A block of layers holds one stored object, so a token
admits exactly one decision, and Section~\ref{sec:exp_evict} prices that single decision against the
strongest per-layer control we could build at a matched byte budget.

\subsection{Sharing the cache across layers}
\label{app:rel_share}

The third line changes the stored object itself, and it appears first in architectures that are trained
to share. Cross-layer attention \citep{brandon2024reducingtransformerkeyvaluecache} lets a group of
layers read one cache, YOCO \citep{sun2024cacheoncedecoderdecoderarchitectures} caches once for an entire
decoder, MLKV \citep{zuhri2024mlkvmultilayerkeyvalueheads} shares key and value heads across layers, and
LCKV \citep{wu2024layercondensedkvcacheefficient} condenses the caches of many layers into one. Together
they establish that a block of layers can live with a single stored object, but they reach it by
retraining the attention stack on billions of tokens, which is not available to a checkpoint that is
already deployed. The same holds along the head axis, where GQA
\citep{ainslie2023gqatraininggeneralizedmultiquery} and multi-head latent attention
\citep{deepseekai2024deepseekv2strongeconomicalefficient} are architectural decisions taken before
training, although recent work converts a trained model to latent attention after the fact
\citep{meng2025transmla, ji2025economicalinferenceenablingdeepseeks}.

Training-free cross-layer compression is the direction this paper works in, and it is recent. xKV
\citep{chang2026xkvcrosslayerkvcachecompression} factorizes the caches of a window of layers into one
shared low-rank basis that it recomputes for every sequence at prefill. CommonKV
\citep{wang2025commonkvcompressingkvcache} derives a shared projection from the grouped layers' own
weight matrices and then merges the token slots whose latents agree across the group. MiniCache
\citep{liu2024minicachekvcachecompression} merges adjacent layers by interpolating per-token unit
vectors while each layer keeps its own norm, KVSharer \citep{yang2024kvsharer} searches for the layer
pairs whose caches are most dissimilar and lets the later layer reuse the earlier one, and XQuant
\citep{yang2025xquantcrosslayer} reuses quantization codes across layers at very low bit-widths. The
single-layer limit of the same construction is a per-layer low-rank cache, which Palu
\citep{chang2024palucompressingkvcachelowrank} and Eigen Attention
\citep{saxena2024eigenattentionattentionlowrank} build on the activation-aware whitening of SVD-LLM
\citep{wang2024svdllm}, and Section~\ref{sec:construction} shows that our objective reduces to exactly
that construction once the cross-layer blocks of the second-moment matrix are set to zero. A recent
position paper argues that quantization dominates rank reduction at any matched budget
\citep{salfati2026quantizationdominatesrankreduction}. Our measurements agree with it in the sense that
matters here, because the axis that pays is the number of layers one stored object serves and not the
rank at which that object is stored.

\section{Extended method}
\label{app:method}

Main Section~\ref{sec:method} states the method; this appendix carries the proof, the gauge
analysis, the full whisper and decode detail, the accounting, and the negative results.

\subsection{Proposition 1: optimality of the calibrated construction}
\label{app:prop1}

The objective of Eq.~(\ref{eq:objective}) is a reduced-rank regression
\citep{izenman1975reducedrank} with the error measured on the keys rather than on the weights.

\begin{quote}
\textbf{Proposition 1.} \emph{Let $\mathbf{G}=\mathbb{E}[\mathbf{H}^{\top}\mathbf{H}]\succ 0$
(enforced by a small ridge $\mathbf{G}\leftarrow\mathbf{G}+\varepsilon\bar{g}\mathbf{I}$, as in
\citealp{wang2024svdllm}), let $\mathbf{M}:=\mathbf{G}^{1/2}\mathbf{W}$ have thin singular value
decomposition $\mathbf{M}=\mathbf{U}\boldsymbol{\Sigma}\mathbf{V}^{\top}$, and let
$R\le\mathrm{rank}(\mathbf{M})$. The minimizers of Eq.~(\ref{eq:objective}) are exactly
$\mathbf{W}_{\downarrow}^{\star}=\mathbf{G}^{-1/2}\mathbf{U}_{:R}\mathbf{T}$ and
$\mathbf{W}_{\uparrow}^{\star}=\mathbf{T}^{-1}\boldsymbol{\Sigma}_{:R}\mathbf{V}_{:R}^{\top}$
for any invertible $\mathbf{T}\in\mathbb{R}^{R\times R}$, with minimum value
$\sum_{i>R}\sigma_i^{2}$.}
\end{quote}

\textbf{Proof.} (1) For any $\mathbf{X}$,
$\mathbb{E}\|\mathbf{H}\mathbf{X}\|_F^{2}
=\mathrm{tr}(\mathbf{X}^{\top}\mathbf{G}\mathbf{X})=\|\mathbf{G}^{1/2}\mathbf{X}\|_F^{2}$;
with $\mathbf{X}=\mathbf{W}-\mathbf{W}_{\downarrow}\mathbf{W}_{\uparrow}$ the objective becomes
the deterministic
$\|\mathbf{G}^{1/2}(\mathbf{W}-\mathbf{W}_{\downarrow}\mathbf{W}_{\uparrow})\|_F^{2}$.
(2) $\mathbf{G}^{1/2}$ is invertible; substituting
$\mathbf{A}:=\mathbf{G}^{1/2}\mathbf{W}_{\downarrow}$ gives
$J=\|\mathbf{M}-\mathbf{A}\mathbf{W}_{\uparrow}\|_F^{2}$.
(3) $\mathbf{A}\mathbf{W}_{\uparrow}$ ranges over exactly the matrices of rank at most $R$, so
the problem is unconstrained best rank-$R$ approximation of $\mathbf{M}$; by the
Eckart--Young--Mirsky theorem \citep{eckart1936approximation, mirsky1960symmetric} the minimizer
is $\mathbf{U}_{:R}\boldsymbol{\Sigma}_{:R}\mathbf{V}_{:R}^{\top}$ with residual
$\sum_{i>R}\sigma_i^{2}$.
(4) Any rank-$R$ factorization of that minimizer is unique up to an invertible right action
$\mathbf{T}$ on $\mathbf{A}$; un-whitening gives the stated pair. \hfill$\blacksquare$

Setting the cross-layer blocks of $\mathbf{G}$ to zero decouples the problem into $n$ per-layer
regressions, the per-layer channel-axis construction of \citet{zhang2026spectra}, which rests on
the activation-aware whitening of \citet{wang2024svdllm}. The primitive is therefore not ours;
the off-diagonal blocks
$\mathbf{G}_{ij}=\mathbb{E}[\mathbf{h}_{\ell_i}\mathbf{h}_{\ell_j}^{\top}]$ are the added,
load-bearing ingredient, and a per-layer method structurally cannot use them. Values use an
independent pair $(\mathbf{W}_{\downarrow}^{V},\mathbf{W}_{\uparrow}^{V})$ obtained identically;
biases are added after reconstruction.

\subsection{The gauge, and why the shipped split is balanced}
\label{app:gauge}

$\mathbf{T}$ is invisible at full precision, since every choice spans the same subspace at the
same minimum, and matters only once the latent is quantized, because it reshapes the latent's
coordinate statistics. Two named choices. \emph{Whiten-flat} ($\mathbf{T}=\mathbf{I}$): all
singular-value energy sits in the fixed up-projection and the cached latent is exactly
isotropic, $\mathbb{E}[\mathbf{c}\mathbf{c}^{\top}]
=\mathbf{U}_{:R}^{\top}\mathbf{G}^{-1/2}\mathbf{G}\mathbf{G}^{-1/2}\mathbf{U}_{:R}=\mathbf{I}_R$.
\emph{Balanced} ($\mathbf{T}=\boldsymbol{\Sigma}_{:R}^{1/2}$; shipped, and also the split of
\citealp{wang2025commonkvcompressingkvcache}):
$\mathbb{E}[\mathbf{c}\mathbf{c}^{\top}]=\boldsymbol{\Sigma}_{:R}$, an energy-ordered latent.

The energy-ordered latent looks like what uniform quantization fears, coordinate variances
spanning orders of magnitude, but a shared scale is defeated by \emph{within-group}
heterogeneity, not by a smooth global slope. The quantizer shares one scale per 128 adjacent
coordinates; in the energy-ordered latent the spectrum decays smoothly, so each group is
internally homogeneous and the per-group min--max scales absorb the ordering across groups:
ordering plus grouping acts as a free piecewise variance-adaptive quantizer. Measured at the
shipped operating point the balanced split is ahead of whiten-flat by a small margin (about
$1.7$ retention points on average), and we keep whiten-flat as an ablation arm. Two honesty
notes. The margin is small, and BellowsKV's low-bit robustness traces to the fixed
activation-weighted basis, not to $\mathbf{T}$. And the isotropy identity stays an analysis
device: equal-variance coordinates would matter under single-scale quantization, which
BellowsKV does not use.

\subsection{Operating points and accounting}
\label{app:accounting}

Uniform $b$-bit quantization is applied per token, per 128-channel group, asymmetric min--max
\citep{liu2024kivi}, with a $0.95$ clip on each group's range at the high tier only (it hurts
below, Appendix~\ref{app:negatives}). The rank stays full ($R{=}p$) at both tiers; a remainder
block shorter than $n$ is kept, never left uncompressed.

\begin{center}
\small
\renewcommand{\arraystretch}{1.13}
\begin{tabular}{lcccc}
\toprule
tier & block & bits & clip & per-token-per-block cost (bits) \\
\midrule
low ($\sim$4$\times$)   & $n{=}2$ & INT8 & 1.0  & $(8R+\tfrac{R}{4}+512)+(8R+\tfrac{R}{4})$ \\
high ($\sim$16$\times$) & $n{=}4$ & INT4 & 0.95 & $(4R+\tfrac{R}{4}+512)+(4R+\tfrac{R}{4})$ \\
top ($\sim$18$\times$)  & $n{=}4$ & K3V2 & 0.95 & $(3R+\tfrac{R}{4}+512)+(2R+\tfrac{R}{4})$ \\
\bottomrule
\end{tabular}
\end{center}

The two brackets are the key and value sides; $R/4$ is each side's fp16 group scale-and-zero,
and $512$ is the key side's whisper coefficients ($r{=}32$ at fp16,
Appendix~\ref{app:whisper}). Every reported ratio counts all of it plus the fp16 window of
Appendix~\ref{app:window}. Measured at 32k context: $3.64\times$ and $13.84\times$ on Llama and
Mistral, $3.53\times$ and $13.13\times$ on Qwen (KV width 512); the high tier at INT3 measures
$17.65\times$ ($16.52\times$ on Qwen) and at K3V2, $(3R+\tfrac{R}{4}+512)+(2R+\tfrac{R}{4})$,
$20.47\times$ ($18.96\times$ on Qwen). Under Protocol A's shorter
mixed-length inputs the window's fixed cost weighs more and the measured ratios are
correspondingly lower; the Protocol-A tables state their own values. Dropping the whisper and
window terms recovers the bare formula $2(bR+R/4)$.

\subsection{The eviction verdict}
\label{app:evict}

BellowsKV-Evict is the block-level form of attention-mass scoring
\citep{zhang2023h2oheavyhitteroracleefficient, li2024snapkvllmknowslooking}; every setting
below is fixed once and used unchanged on all three models.

\textbf{Scoring.} After prefill, the queries of the last $64$ prompt positions attend to every
cached position, and each cached token receives the attention mass these queries place on it,
summed over the block's $n$ layers and all their heads. The scores are computed on the keys the
block will actually read at decode, reconstructed from the quantized latent through the
read-heads and whisper-corrected, not on the exact prefill keys, so the verdict sees exactly
what decode will see. The summed profile is then max-pooled over a width-$13$ window along the
sequence, so that a token kept for its own score also keeps its neighbours
\citep{li2024snapkvllmknowslooking}.

\textbf{Selection.} The first $4$ positions and the last $64$ prompt positions are exempt and
always kept (Appendix~\ref{app:window}). Of the remaining positions, the fraction $q$ with the
highest pooled scores survives; the rest are dropped once for the whole block, a single gather
applied to the latent and, in the composed operator, to its group scales and whisper
coefficients alike. Because the block holds one slot per token, the gather is issued once per
block rather than once per layer, so the verdict costs $1/n$ of its per-layer counterpart and is
consistent across the block by construction.

\textbf{Accounting.} The keep ratio $q$ is applied to the compressed cache, so the stored size
per block is $q$ times the per-token cost of Appendix~\ref{app:accounting} plus the exempt
window. BellowsE alone prunes the $n{=}4$ fp16 latent; BellowsQ$\times$BellowsE prunes the
INT4 tier. At the paired points of Table~\ref{app:evict_full} BellowsE stores $1$--$5\%$ more
than the per-layer control at the same nominal budget, which is the direction that disfavours
it.

\textbf{The per-layer control.} The control runs the identical scoring and pooling, but each
layer scores on its own exact fp16 cache and keeps its own top fraction, so its verdicts differ
layer to layer and its gather is issued $n$ times. The naive stack of Section~\ref{sec:exp_evict}
(xKV + INT4 + SnapKV) needs a verdict pooled per xKV group for the same reason BellowsE does:
per-layer verdicts cannot share one stored basis.

\subsection{The recency window}
\label{app:window}

The first $4$ positions and the last $64$ prompt positions keep their original per-layer fp16
tensors, and decode-time tokens are appended per-layer as usual, so the recent region is exact
at all times: the standard residual-window practice \citep{liu2024kivi, xiao2024streamingllm}
applied at the block level. Cost: $68$ prompt positions, $0.2\%$ of a 32k context (charged in
Appendix~\ref{app:accounting}). What it buys concentrates where the codec is weakest:
recency-dominated tasks such as code completion and long-form generation recover $5$--$11$ raw
points at the high tier, while retrieval-style tasks are unchanged.

\subsection{The whisper in full}
\label{app:whisper}

\textbf{Construction.} At prefill the true keys are still in hand, so the residual
$\mathbf{E}=\mathbf{K}-\hat{\mathbf{K}}$ (pre-rotation, bias-free) is free to read. A randomized
range finder \citep{halko2010findingstructurerandomnessprobabilistic} sketches its top $r{=}32$
right-singular directions: draw a Gaussian
$\boldsymbol{\Omega}\in\mathbb{R}^{np\times(r+8)}$, form $\mathbf{E}\boldsymbol{\Omega}$,
orthonormalize, project, and take the top-$r$ right singular vectors $\mathbf{V}_r$; three
matrix products and one tiny SVD, milliseconds per block. Stored: $\mathbf{V}_r$ once per
sequence ($np\times 32$ at fp16, about $0.7\%$ of the cache at 32k) and $32$ fp16 coefficients
per token (the $512$ bits of Appendix~\ref{app:accounting}). Every later key read adds back
$\mathbf{C}_t\mathbf{V}_r$, one rank-$32$ product. Measured, $\|\mathbf{E}\|$ is roughly
$10\%$ of $\|\mathbf{K}\|$ and the sketch captures about three quarters of it.

\textbf{What it buys.} On averages the whisper is nearly free-standing: removing it costs about
$0.1$ mean retention points at the high tier ($97.9$ against $98.0$, Llama, Protocol B,
Table~\ref{app:tab:ablation}), and at the fp16 sharing-only points the paired no-whisper arms
sit $0.05$--$0.4$ points lower. Its value is in the tail: the largest per-task swing is $4$
points (2WikiMQA, $101.5$ with the whisper against $97.5$ without). The whisper is a worst-case
stabilizer, not an average lifter, and we claim no more for it.

\textbf{Keys only, and why.} A key error passes through the softmax and perturbs \emph{which}
positions are attended, a brittle failure worth correcting everywhere. A value error enters the
convex combination $\sum_t a_t\mathbf{v}_t$ and averages out over a long prefill. Patching
values helps exactly when the prefill is long (the $+$V-whisper row of
Table~\ref{app:tab:ablation}: $98.7$ against $98.0$ at 32k) and flips sign on short-prefill,
generation-heavy inputs, where the rank-$32$ structure perturbs the few vectors that dominate
the average. One rule is shipped uniformly: whisper the keys, leave the values to the attention
average.

\subsection{Decode paths}
\label{app:decode}

\textbf{Mode A (shipped): values absorbed, keys reconstructed.} Values carry no positional
rotation, so the read-head commutes with the attention-weighted sum,
\begin{equation}
\sum\nolimits_{t} a_{t}\,\big(\mathbf{W}_{\uparrow}^{V,(i)\,\top}\mathbf{c}^{V}_{t}\big)
=\mathbf{W}_{\uparrow}^{V,(i)\,\top}\Big(\sum\nolimits_{t} a_{t}\,\mathbf{c}^{V}_{t}\Big),
\label{eq:absorption}
\end{equation}
the weight-absorption identity of multi-head latent attention
\citep{deepseekai2024deepseekv2strongeconomicalefficient}: every layer attends directly in the
$R$-dimensional latent, $\mathbf{W}_{\uparrow}^{V,(i)}$ folds into the output projection, and
the value path is exact up to the codec itself. Keys cannot use the identity because the
positional rotation sits between the latent and the score: each key is reconstructed through
its read-head, whisper-corrected, rotated per head, and attended, a rank-$R$ plus rank-$32$
product per layer that hides under the memory-bound cache read.

\begin{table}[H]
\centering
\caption{\textbf{BellowsKV decodes at parity, and the same relation prices Palu's speedup within 6\%.}
Decode speed and capacity (H100 SXM, HuggingFace Transformers, 32k context, no custom kernel): the
reconstruct-at-read relation of Section~\ref{sec:exp_decode} on both designs it covers, and what the freed
bytes buy under two engines.}
\label{app:tab:decode}
\vspace{4pt}
\small
\renewcommand{\arraystretch}{1.05}
\begin{tabular}{lccc}
\toprule
Design & Read path & Relation predicts & Measured / reported \\
\midrule
\multicolumn{4}{c}{\textit{decode speed}} \\
\midrule
BellowsKV & $R{=}d_{kv}$, fp16 & 1.00$\times$ & 0.995$\times$ (32.5 tok/s) \\
Palu (rank-cutting) & $R{=}d_{kv}/2$, fp16 & 2.00$\times$ & 1.89$\times$ (reported) \\
\midrule
\multicolumn{4}{c}{\textit{capacity from the freed bytes}} \\
\midrule
HuggingFace Transformers & max batch & -- & 2$\times$ \\
vLLM paged allocator & KV pool & -- & 14.76$\times$ (6.02M tokens) \\
\bottomrule
\end{tabular}
\vspace{3pt}
\end{table}

\textbf{Mode B (optional restructuring, not a speedup).} An MLA-style decoupled-rotation split
of the key path \citep{deepseekai2024deepseekv2strongeconomicalefficient, meng2025transmla,
ji2025economicalinferenceenablingdeepseeks} can move the score into the latent space. The
relation of Section~\ref{sec:exp_decode} shows why we do not ship it: at $R=d_{kv}$ consumed in
fp16 both of its factors are one, so absorption removes arithmetic but not bytes, and measured
decode is $0.995\times$ dense either way. Only a smaller $R$ or a sub-fp16 consumption format
changes the bound, and those are exactly the moves the accuracy story forbids or defers.

\subsection{Costs}
\label{app:costs}

Calibration is one-shot: one forward pass over $32$ windows of $2048$ tokens (half WikiText-2,
half code) accumulates the second-moment blocks, then one singular value decomposition per
block; about twenty minutes on one GPU per model, no gradients, and the factors fold into the
weights. We fix this one corpus recipe globally rather than tuning the mix per model or task.
Per sequence, the only added computation is the whisper sketch (milliseconds per block, roughly
two orders of magnitude below xKV's per-sequence full SVD; a fully per-sequence variant of our
own construction costs $2.2\times$ the prefill and scores worse,
Appendix~\ref{app:negatives}). Decode adds one dequantization per block and the products of
Appendix~\ref{app:decode}.

\subsection{Negative results}
\label{app:negatives}

Each of these was implemented, run, and rejected; they bound what the ingredients do and do not
buy.

\begin{enumerate}
\item \textbf{Sharing reconstructed keys across layers.} Token-wise cosine between
adjacent-layer keys is near zero ($-0.002$ on Qwen, $+0.0016$ on Llama), and substituting one
shared key across even two layers inflates perplexity by three orders of magnitude under an
oracle choice of the shared key (diagnostic probe). Sharing must happen one step earlier, in
the latent; this is the design fact behind Section~\ref{sec:ob1}.
\item \textbf{Quantization-aware read-head refit} (ridge refit of $\mathbf{W}_{\uparrow}$
against the quantized latent): flat to negative on probe cells. Proposition~1's solution is
already near-optimal, and INT4 noise on the latent is effectively unbiased for the objective.
\item \textbf{Per-sequence gauge} (whitening the latent per sequence around the quantizer):
$96.7$ against the shipped $98.0$ mean retention. The per-group scales already deliver the
adaptation, and the sequence-level estimate adds variance.
\item \textbf{Fully per-sequence basis} (this sequence's optimal rank-$R$ SVD in place of the
calibrated consensus): $97.3$ at $2.2\times$ prefill cost, below the frozen consensus.
Test-time subspace adaptation is not the active ingredient; a stable average basis plus a
rank-$32$ correction beats a fully sequence-specific basis.
\item \textbf{V-whisper as a default}: the sign flips with prefill length
(Appendix~\ref{app:whisper}); an ablation row, not the method.
\item \textbf{Whiten-flat split as the shipped gauge}: behind balanced by about $1.7$ points at
the operating point (Appendix~\ref{app:gauge}); kept as the ablation arm.
\item \textbf{Clip below the high tier}: the $0.95$ range clip helps only at
$n{=}4\times$INT4; at INT8 it hurts and is disabled.
\item \textbf{Depth-adaptive rank allocation and non-adjacent groupings}: both flat to negative
in probes; consecutive blocks at uniform full rank remain the default.
\end{enumerate}

\section{Extended experiments}
\label{app:extended_exp}

\subsection{Setup, accounting, and coverage}
\label{app:exp_setup}

\paragraph{The two protocols.} Every experiment runs under one of two protocols, one per
baseline family, and no number crosses between them.

\begin{center}
\small
\renewcommand{\arraystretch}{1.2}
\begin{tabular}{p{0.16\textwidth}p{0.36\textwidth}p{0.36\textwidth}}
\toprule
 & \textbf{Protocol A} & \textbf{Protocol B} \\
\midrule
question & absolute accuracy under compression & fraction of dense accuracy retained \\
harness & official LongBench pipeline & the harness released by the xKV authors \\
inputs & 8 English tasks, mixed lengths & 11 tasks, only inputs longer than 4k tokens (up to the 32k window) \\
unit & score 0--100 & retention \% = score $\div$ dense score \\
compared against & the quantization family & the cross-layer and eviction families \\
sample & full test sets (200/500 per task) & all qualifying samples (Table~\ref{app:tab:pbcounts}) \\
\bottomrule
\end{tabular}
\end{center}

Retention is computed per task against the dense run of the same harness and campaign; values
above 100\% occur and are expected. Task codes: Qsp (Qasper), MFQ (MultiFieldQA-en), HQA (HotpotQA), MSQ (MuSiQue),
NQA (NarrativeQA), Gov (GovReport), QMS (QMSum), PsR (Passage Retrieval-en), LCC, RBP
(RepoBench-P), 2Wk (2WikiMultihopQA).

\begin{table}[H]
\centering
\caption{\textbf{Protocol B qualifying sample per task} (inputs longer than 4k tokens and
within the 32k window; the count differs per model because tokenizers differ). Every Protocol-B
number in the paper uses all qualifying samples.}
\label{app:tab:pbcounts}
\vspace{7pt}
\scriptsize
\renewcommand{\arraystretch}{1.13}
\setlength{\tabcolsep}{3.2pt}
\begin{tabular}{lccccccccccc}
\toprule
Model & Qsp & MFQ & HQA & MSQ & NQA & Gov & QMS & PsR & LCC & RBP & 2Wk \\
\midrule
Llama-3.1-8B & 122 & 110 & 195 & 200 & 115 & 177 & 199 & 200 & 93 & 442 & 175 \\
Mistral-7B & 143 & 117 & 195 & 200 & 85 & 182 & 190 & 200 & 150 & 470 & 183 \\
Qwen2.5-7B & 125 & 111 & 195 & 200 & 115 & 180 & 199 & 200 & 96 & 444 & 178 \\
\bottomrule
\end{tabular}
\vspace{3pt}
\end{table}

\paragraph{BellowsKV configuration.} One codec serves every table: balanced split, $32$
calibration windows, full rank, sink $4$ + recent $64$, key whisper $r{=}32$. BellowsQ ships
two tiers, $n{=}2$ INT8 (no clip) and $n{=}4$ INT4 (clip $0.95$); the high tier is also stored at
INT3 and at K3V2 (key latent 3 bits, value latent 2 bits) with the same codec and clip, and the
$\sim$8$\times$ and $\sim$12$\times$ appendix tiers use $n{=}2$ and $n{=}3$ at INT4. BellowsE is the $n{=}4$ fp16
latent pruned by the verdict of Section~\ref{sec:decode}; BellowsS is that latent with neither
operator. The accounting formula and the measured per-model ratios are in
Appendix~\ref{app:accounting}; every reported ratio counts all metadata.

\paragraph{Compression tiers.} Every comparison table groups its rows into tiers of \emph{measured}
compression ratio (Table~\ref{app:tab:tiers}), sorts a tier by ratio, and marks its best and second
average. A row belongs to the tier its own measured ratio falls in, never to a nominal one, so
two rows between the same pair of rules are the rows to compare, and the Comp.\ column shows
how close their bytes are. The bins are fixed once for each axis and used unchanged in the
main text and the appendix.

\begin{table}[H]
\centering
\caption{\textbf{Compression tiers.} Bins on the measured ratio (all metadata counted) that every comparison table uses; a row belongs to the tier its own measured ratio falls in, never to a nominal one. The edges are chosen so that the highest ratio inside a tier is always a BellowsKV point, which puts every within-tier comparison on the baselines' side of the byte budget, and so that no tier holds our rows alone. Protocol A's top tier is the one exception: no per-layer quantizer exists past $8\times$.}
\label{app:tab:tiers}
\vspace{5pt}
\small
\renewcommand{\arraystretch}{1.12}
\begin{tabular}{@{}lll@{}}
\toprule
Axis & Tier & Measured ratio \\
\midrule
Protocol A (quantizers) & $<$4$\times$ & $<$4$\times$ \\
 & $\sim$5$\times$ & 4--6.6$\times$ \\
 & $\sim$8$\times$ & 6.6--9.5$\times$ \\
 & $\geq$12$\times$ & $\geq$9.5$\times$ (no per-layer point exists) \\
\addlinespace[2pt]
Protocol B (sharing) & $\leq$4$\times$ & 1--3.9$\times$ (as-shipped fp16 baselines) \\
 & 4--11$\times$ & 3.9--10.9$\times$ \\
 & $\sim$12$\times$ & 10.9--14$\times$ \\
 & $\geq$14$\times$ & 14--24$\times$ \\
\addlinespace[2pt]
Eviction (budget) & $\sim$2$\times$ & 1.5--3$\times$ \\
 & $\sim$4$\times$ & 3--6$\times$ \\
 & $\sim$8$\times$ & 6--10$\times$ \\
 & $\sim$16$\times$ & 13--20$\times$ \\
 & $\sim$30$\times$ & 25--35$\times$ \\
 & $\sim$57$\times$ & 50--60$\times$ \\
\addlinespace[2pt]
\bottomrule
\end{tabular}
\vspace{3pt}
\end{table}

\paragraph{Baseline configurations.} Tables~\ref{app:tab:cfg_a} and~\ref{app:tab:cfg_b} give every baseline
configuration behind the numbers in this paper: the Protocol-A quantizers at their published settings, and the
Protocol-B cross-layer methods at the settings that produce each baseline row of Table~\ref{tab:protoB}.
Protocol-A baselines run each method's own official code, with group size $128$ and each method's own
residual-window convention. Protocol-B baselines all run inside the xKV harness on the same qualifying samples
against the same dense run.

\begin{table}[H]
\centering
\caption{\textbf{Every Protocol-A baseline configuration, and the models it exists for.} Each method runs its own official code at its published settings, with group size $128$ and the method's own residual-window convention. $\checkmark$ marks a model the method ships modelling for; coverage on Qwen is thinner than on the other two, which is stated wherever it matters. $^\dagger$RotateKV's 4-bit configuration exists only in the Mistral config set. $^\ddagger$Neither PolarQuant nor TurboQuant ships a Qwen path, so the Qwen adapters are ours and quantize bias-free keys (see \emph{Compatibility fixes}).}
\label{app:tab:cfg_a}
\vspace{4pt}
\small
\renewcommand{\arraystretch}{1.12}
\setlength{\tabcolsep}{5pt}
\begin{tabular}{@{}llccc@{}}
\toprule
Method & Configurations run & Llama & Mistral & Qwen \\
\midrule
KIVI & K4V4, K2V2 & \checkmark & \checkmark & -- \\
OTT & K4V4, K2V2 & \checkmark & \checkmark & -- \\
RotateKV & 4-bit$^\dagger$, 3-bit, 2-bit & \checkmark & \checkmark & -- \\
TurboQuant & K4V4, K4V2, K2V2 & \checkmark & \checkmark & \checkmark$^\ddagger$ \\
PolarQuant & r4t4v4, r3t3v4, r2t2v4, r2t2v2 & \checkmark & -- & \checkmark$^\ddagger$ \\
\bottomrule
\end{tabular}
\vspace{2pt}
\end{table}

\begin{table}[H]
\centering
\caption{\textbf{The setting behind every Protocol-B baseline row of Table~\ref{tab:protoB}, with the ratio it measures.} All of them run inside the xKV harness on the same qualifying samples against the same dense run; the four training-free methods other than xKV were ported into it for this paper and the harness patches are released with the code. Where a setting differs on Qwen, whose KV width is half and which has $28$ layers, the Qwen value is in parentheses; where all three differ the values are listed Llama, Mistral, Qwen. $p$ is xKV's full per-sequence rank. Storage bits are the method's own composition, not ours.}
\label{app:tab:cfg_b}
\vspace{4pt}
\small
\renewcommand{\arraystretch}{1.12}
\setlength{\tabcolsep}{5pt}
\begin{tabular}{@{}llccc@{}}
\toprule
 & & \multicolumn{3}{c}{measured compression} \\
\cmidrule(l){3-5}
Method & Setting & Llama & Mistral & Qwen \\
\midrule
xKV & rank $p$, fp16 & $4.00\times$ & $4.00\times$ & $4.00\times$ \\
xKV & rank $p/2$, fp16 & -- & -- & $8.00\times$ \\
xKV & rank $p/4$, fp16 & $16.00\times$ & $16.00\times$ & $16.00\times$ \\
\addlinespace[2pt]
CommonKV & $r{=}3328$ ($r{=}1664$), INT4 & $3.95\times$ & $3.95\times$ & $3.95\times$ \\
CommonKV & $r{=}1664$ ($r{=}832$), INT4 & $7.89\times$ & $7.90\times$ & $7.89\times$ \\
CommonKV & $r{=}832$ ($r{=}416$), INT4 & $15.78\times$ & $15.80\times$ & $15.78\times$ \\
\addlinespace[2pt]
MiniCache fp16 & deeper half, fp16 & $1.31\times$ & $1.31\times$ & $1.31\times$ \\
MiniCache+INT4 & deeper half, INT4 & $4.78\times$ & $4.78\times$ & $4.78\times$ \\
MiniCache all+INT4 & every pair, INT4 & $6.72\times$ & $6.72\times$ & $6.72\times$ \\
MiniCache+INT2 & every pair, INT2 & $11.96\times$ & $11.99\times$ & $11.97\times$ \\
\addlinespace[2pt]
KVSharer fp16 & $8/32$ shared ($7/28$), fp16 & $1.33\times$ & $1.33\times$ & $1.33\times$ \\
KVSharer+INT4 & $8/32$ shared ($7/28$), INT4 & $4.99\times$ & $4.99\times$ & $4.99\times$ \\
KVSharer 37.5\%+INT2 & $12/32$ shared, INT2 & -- & $11.24\times$ & -- \\
KVSharer+INT2 & $12/32$ / $16/32$ / $10/28$ shared, INT2 & $11.24\times$ & $14.05\times$ & $10.93\times$ \\
\addlinespace[2pt]
XQuant published & classical schedule & $6.59\times$ & $6.59\times$ & $6.59\times$ \\
XQuant 1-bit & 1-bit everywhere, groups $64$ & $15.11\times$ & $15.11\times$ & $15.11\times$ \\
\bottomrule
\end{tabular}
\vspace{2pt}
\end{table}

\paragraph{Compatibility fixes.} Three baselines needed a change before they would run at all, and in each case
the change removes a handicap rather than adding one. PolarQuant's released code ships only its key-side path,
which leaves values in fp16 at about $1.6\times$, with the Mixed K+V path present but commented out; we enable
it, because the key-only path cannot reach the operating points its own paper reports. On Qwen, PolarQuant and
TurboQuant quantize bias-free keys, that is $k$ minus the rotated $k$-projection bias, added back at read time;
the bias is a weight, so the ratio is unchanged. Qwen2.5's key bias dominates the key norm, and both methods
code direction and magnitude, so on raw keys they collapse and score higher at fewer bits; those raw-key runs
are kept invalidated and are not reported. Neither method ships a Qwen path, so the Qwen adapters are ours, as
are CommonKV's Qwen merge weights, which we recomputed with the authors' own sensitivity procedure, the
perplexity increase when one layer's keys are randomized, normalized within each group; it reproduces their
shipped Llama list to a correlation of $0.99$. MiniCache released no code, and its printed inequality would
retain the least divergent token rows rather than the most divergent ones, so we follow its prose and its
$5\%$ budget.

\paragraph{Accounting.} Two baselines are charged differently from their papers, in both cases because the
published figure omits bytes the method actually stores. CommonKV's official code installs the same
up-projection for keys and values, so a layer's key latent and value latent are one $r$-dimensional vector: we
store and charge it once, and we credit its cross-layer token merge, under which the $40\%$ of token slots
whose first- and last-layer latents are least similar stay per-layer, together with the first $4$ and last
$128$ positions, while every other slot holds one weighted average per group. Its rank is then set so that this
honest ratio lands on the tier. XQuant's published ``1.375-bit'' figure excludes scales and zeros, which at
group $32$ add a full bit per value; our ratio counts them and the $128$-token residual.

\paragraph{Coverage.} Protocol-A coverage is uneven because KIVI, OTT and RotateKV ship no Qwen modelling and
PolarQuant ships no Mistral modelling; Table~\ref{app:tab:cfg_a} states which rows exist, and Qwen's
Protocol-A field is thinner than the other two models' wherever it matters. On the Protocol-B side, xKV's Qwen
runs at rank $p$ and $p/2$ lack the MuSiQue cell, so their averages are over ten tasks and carry a $^\dagger$
wherever they appear. KVSharer's Qwen rows are low at every setting because its own strategy search makes layer
$0$ the donor for layers $1$--$3$, its most dissimilar pairs, and the model does not tolerate that at long
inputs. For every Protocol-B row, sharing is applied to the cache read at decode and prefill attention is
exact.

\begin{figure}[H]
  \centering
  \includegraphics[width=\linewidth]{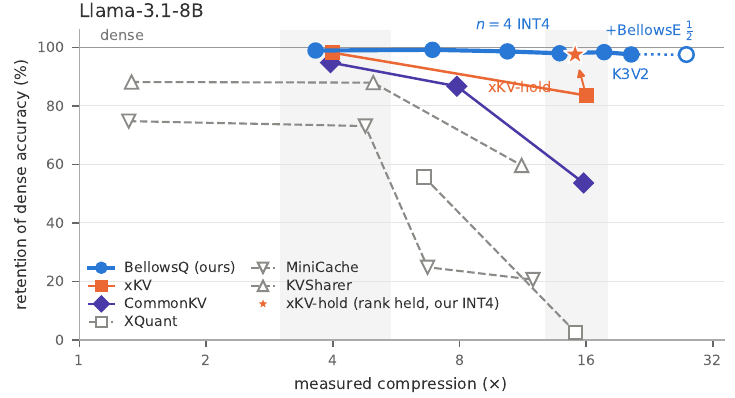}
  \caption{\textbf{On Llama the cross-layer axis stalls at $16\times$ because rank is cut, and holding the rank
  recovers it.} Protocol~B retention of dense accuracy against measured compression, one line per training-free
  cross-layer method over all of its registered operating points, one harness, full sample. Filled markers are the
  latent methods and hollow grey markers the raw-cache methods; the shaded bands are the $\sim$4$\times$ and
  $\sim$16$\times$ tiers of Table~\ref{tab:protoB}. The star is xKV-hold, xKV's own basis kept at near-full rank and
  stored with our INT4 leg instead of truncated (Section~\ref{sec:exp_rank}); the hollow blue circle is BellowsQ
  composed with a keep-$1/2$ verdict (Section~\ref{sec:exp_evict}). $^\dagger$10 tasks.}
  \label{fig:pareto_llama}
\end{figure}

\begin{figure}[H]
  \centering
  \includegraphics[width=\linewidth]{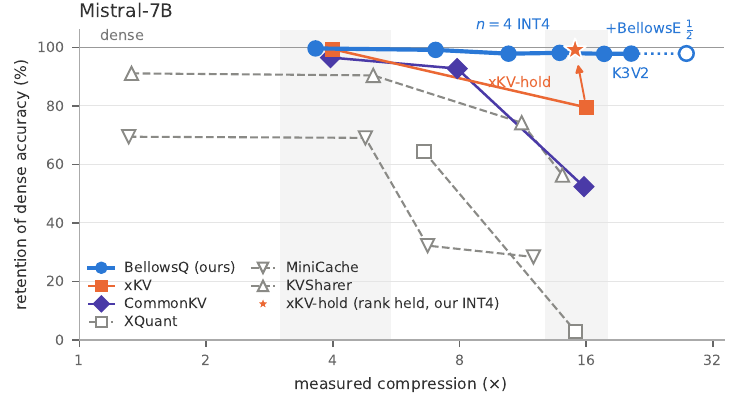}
  \caption{\textbf{Mistral repeats the pattern: every latent method works at $\sim$4$\times$, only the full-rank
  latent survives past $12\times$, and the transplant recovers xKV to within a point of the consensus.} Axes, series
  and markers as in Figure~\ref{fig:pareto_llama}.}
  \label{fig:pareto_mistral}
\end{figure}

\begin{figure}[H]
  \centering
  \includegraphics[width=\linewidth]{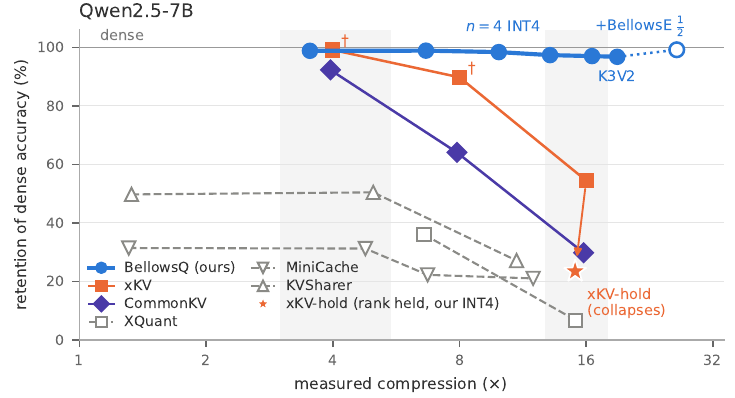}
  \caption{\textbf{Qwen is where the rivals fall furthest and where the transplant collapses.} xKV loses more than
  forty points between $4\times$ and $16\times$, and xKV-hold, which recovers on the other two models, drops to
  $23.4\%$ here, for a cause we have not isolated (Section~\ref{sec:exp_rank}). Axes, series and markers as in
  Figure~\ref{fig:pareto_llama}.}
  \label{fig:pareto_qwen}
\end{figure}

Table~\ref{tab:protoB} in the main text is the complete Protocol-B picture: every operating point of every
method on every model, tier by tier (bins in Table~\ref{app:tab:tiers}), including the baselines'
as-published fp16 settings and every low-bit setting they compose with. KVSharer's largest
shared set with INT2 lands at 14.1$\times$ on Mistral and near 11$\times$ on Llama and Qwen, so the same
setting sits in different tiers on different models; the table follows the measured ratio.
Figures~\ref{fig:pareto_llama}--\ref{fig:pareto_qwen} draw the same rows as a frontier, one figure per
model: retention against measured compression, one line per method.

\paragraph{The CommonKV numbers.} Under the accounting above CommonKV is a working $4\times$ method. An
earlier version of this comparison, taken from its own harness, charged two latents per layer and gave the
merge no credit, which placed its $4\times$ point at a rank where the method has already collapsed. What the
corrected rows then test is whether the axis extends, which is this paper's question, and the mechanism at
$16\times$ is the one Section~\ref{app:ablation_full} prices: the per-layer latent must fall to a rank below the
viability floor before the bytes are saved.

\subsection{Protocol A: full leaderboards}
\label{app:protoA_full}

Tables~\ref{app:tab:lb_llama318b}--\ref{app:tab:lb_qwen257b} give the complete Protocol-A
picture per model, tier by tier: every BellowsKV configuration from $1.92\times$ to $18.6\times$
(BellowsS rows are the sharing-only fp16 latent; BellowsQ rows add uniform bits, and the K3V2
row keeps the key latent at 3 bits while storing the value latent at 2) against every official
baseline configuration. The $\geq$12$\times$ tier holds our rows alone, the one tier in the paper that does: a
per-layer quantizer at full rank bottoms out at 2 bits, which is $8\times$, and going further forces rank
truncation, the regime Section~\ref{app:ablation_full} shows collapsing, so no per-layer point exists there to
compare with. Every other tier, here and in Table~\ref{tab:protoB}, holds at least one baseline and has a
BellowsKV point as its highest ratio (Table~\ref{app:tab:tiers}).

\begin{table}[H]
\centering
\caption{\textbf{Protocol A, Llama-3.1-8B.} Absolute LongBench scores, full test sets, every BellowsKV point and every official baseline configuration, grouped by measured compression tier (rules separate tiers; bins in Table~\ref{app:tab:tiers}); within a tier rows are sorted by ratio and the \colorbox{blue!18}{\textbf{best}} and \colorbox{blue!7}{\underline{second}} Avg are marked. Baselines run each method's official code; PolarQuant runs its Mixed K+V path, and on Qwen it and TurboQuant quantize bias-free keys (Appendix~\ref{app:exp_setup}).}
\label{app:tab:lb_llama318b}
\vspace{5pt}
\fontsize{9pt}{11pt}\selectfont
\renewcommand{\arraystretch}{1.0}
\setlength{\tabcolsep}{5pt}
\resizebox{0.98\linewidth}{!}{%
\begin{tabular}{@{}lrccccccccc@{}}
\toprule
Method & Comp. & Qasper & QMSum & MNews & TREC & TQA & SAMSum & LCC & RB-P & Avg \\
\midrule
Dense fp16 & $1.00\times$ & 46.57 & 25.19 & 27.03 & 72.50 & 91.65 & 43.47 & 63.38 & 56.55 & \textbf{53.29} \\
\midrule
BellowsS $n{=}2$ & $1.95\times$ & 46.76 & 25.32 & 26.08 & 72.50 & 91.28 & 43.90 & 62.38 & 56.32 & 53.07 \\
BellowsS $n{=}3$ & $2.90\times$ & 45.25 & 25.13 & 25.90 & 72.50 & 91.44 & 43.69 & 62.28 & 55.15 & 52.67 \\
PolarQuant r4t4v4 & $3.66\times$ & 46.01 & 25.56 & 26.69 & 72.50 & 91.86 & 42.88 & 65.38 & 58.44 & 53.66 \\
BellowsQ $n{=}2$ INT8 & $3.67\times$ & 46.47 & 25.39 & 26.19 & 72.50 & 91.28 & 43.77 & 62.52 & 56.28 & 53.05 \\
\cellcolor{blue!18}\textbf{KIVI K4V4} & $3.76\times$ & 46.11 & 25.14 & 27.11 & 72.50 & 92.42 & 44.10 & 65.05 & 58.91 & \cellcolor{blue!18}\textbf{53.92} \\
\cellcolor{blue!7}\underline{OTT K4V4} & $3.76\times$ & 46.47 & 25.30 & 26.66 & 72.50 & 91.73 & 43.73 & 64.96 & 59.28 & \cellcolor{blue!7}\underline{53.83} \\
BellowsS $n{=}4$ & $3.84\times$ & 46.30 & 24.97 & 25.26 & 72.50 & 91.49 & 44.70 & 61.23 & 55.80 & 52.78 \\
\midrule
\cellcolor{blue!7}\underline{TurboQuant K4V4} & $4.00\times$ & 45.87 & 25.21 & 26.79 & 73.00 & 91.06 & 43.68 & 62.44 & 54.65 & \cellcolor{blue!7}\underline{52.84} \\
\cellcolor{blue!18}\textbf{PolarQuant r3t3v4} & $4.13\times$ & 45.60 & 25.38 & 26.60 & 72.50 & 92.16 & 42.44 & 65.36 & 57.97 & \cellcolor{blue!18}\textbf{53.50} \\
PolarQuant r2t2v4 & $4.74\times$ & 42.52 & 24.37 & 25.46 & 66.00 & 91.50 & 40.91 & 61.13 & 53.18 & 50.63 \\
RotateKV 3-bit & $5.33\times$ & 45.67 & 24.69 & 26.85 & 73.50 & 90.34 & 42.86 & 61.42 & 53.71 & 52.38 \\
TurboQuant K4V2 & $5.33\times$ & 43.94 & 25.14 & 26.29 & 71.50 & 91.23 & 42.87 & 62.57 & 57.01 & 52.57 \\
BellowsQ $n{=}3$ INT8 & $5.42\times$ & 45.62 & 25.74 & 25.55 & 72.50 & 91.53 & 44.06 & 62.03 & 55.67 & 52.84 \\
\midrule
PolarQuant r2t2v2 & $6.74\times$ & 41.66 & 24.03 & 23.54 & 59.00 & 92.25 & 38.88 & 58.23 & 50.46 & 48.51 \\
\cellcolor{blue!7}\underline{BellowsQ $n{=}2$ INT4} & $6.75\times$ & 46.40 & 24.90 & 26.06 & 72.50 & 91.27 & 44.29 & 61.82 & 56.16 & \cellcolor{blue!7}\underline{52.92} \\
KIVI K2V2 & $7.11\times$ & 42.60 & 24.73 & 26.45 & 71.50 & 91.14 & 43.42 & 61.75 & 55.80 & 52.17 \\
\cellcolor{blue!18}\textbf{OTT K2V2} & $7.11\times$ & 44.98 & 24.81 & 26.61 & 72.50 & 91.85 & 43.95 & 64.25 & 57.44 & \cellcolor{blue!18}\textbf{53.30} \\
TurboQuant K2V2 & $8.00\times$ & 40.50 & 23.17 & 24.63 & 62.00 & 86.96 & 42.82 & 48.80 & 42.57 & 46.43 \\
RotateKV 2-bit & $8.00\times$ & 37.92 & 23.41 & 26.38 & 61.00 & 87.64 & 39.71 & 53.47 & 46.50 & 47.00 \\
BellowsQ $n{=}2$ INT3 & $8.53\times$ & 45.89 & 25.30 & 25.80 & 72.50 & 91.60 & 44.29 & 60.80 & 55.62 & 52.72 \\
\midrule
\cellcolor{blue!18}\textbf{BellowsQ $n{=}4$ INT4} & $12.73\times$ & 45.65 & 24.86 & 25.63 & 72.50 & 91.18 & 44.66 & 60.47 & 54.62 & \cellcolor{blue!18}\textbf{52.45} \\
\cellcolor{blue!7}\underline{BellowsQ $n{=}4$ INT3} & $15.85\times$ & 44.39 & 25.09 & 25.57 & 72.50 & 91.38 & 44.03 & 61.01 & 55.40 & \cellcolor{blue!7}\underline{52.42} \\
BellowsQ $n{=}4$ K3V2 & $18.07\times$ & 44.13 & 25.18 & 25.60 & 72.00 & 90.65 & 44.41 & 60.65 & 55.51 & 52.27 \\
\bottomrule
\end{tabular}
}
\vspace{3pt}
\end{table}

\begin{table}[H]
\centering
\caption{\textbf{Protocol A, Mistral-7B.} Absolute LongBench scores, full test sets, every BellowsKV point and every official baseline configuration, grouped by measured compression tier (rules separate tiers; bins in Table~\ref{app:tab:tiers}); within a tier rows are sorted by ratio and the \colorbox{blue!18}{\textbf{best}} and \colorbox{blue!7}{\underline{second}} Avg are marked. Baselines run each method's official code; PolarQuant runs its Mixed K+V path, and on Qwen it and TurboQuant quantize bias-free keys (Appendix~\ref{app:exp_setup}).}
\label{app:tab:lb_mistral7b}
\vspace{5pt}
\fontsize{9pt}{11pt}\selectfont
\renewcommand{\arraystretch}{1.0}
\setlength{\tabcolsep}{5pt}
\resizebox{0.98\linewidth}{!}{%
\begin{tabular}{@{}lrccccccccc@{}}
\toprule
Method & Comp. & Qasper & QMSum & MNews & TREC & TQA & SAMSum & LCC & RB-P & Avg \\
\midrule
Dense fp16 & $1.00\times$ & 38.15 & 25.57 & 26.43 & 76.00 & 88.59 & 47.46 & 59.46 & 60.56 & \textbf{52.78} \\
\midrule
\cellcolor{blue!7}\underline{BellowsS $n{=}2$} & $1.96\times$ & 38.75 & 25.62 & 26.06 & 76.00 & 88.81 & 46.78 & 60.36 & 61.97 & \cellcolor{blue!7}\underline{53.04} \\
BellowsS $n{=}3$ & $2.91\times$ & 38.20 & 25.04 & 25.75 & 76.00 & 89.57 & 47.17 & 59.84 & 61.41 & 52.87 \\
BellowsQ $n{=}2$ INT8 & $3.69\times$ & 38.49 & 25.34 & 25.81 & 76.00 & 88.56 & 46.74 & 60.42 & 61.99 & 52.92 \\
OTT K4V4 & $3.76\times$ & 37.95 & 25.80 & 26.50 & 76.00 & 88.89 & 47.08 & 60.40 & 61.30 & 52.99 \\
KIVI K4V4 & $3.76\times$ & 36.88 & 25.42 & 26.58 & 76.00 & 88.89 & 47.09 & 60.53 & 60.99 & 52.80 \\
\cellcolor{blue!18}\textbf{BellowsS $n{=}4$} & $3.86\times$ & 37.67 & 25.42 & 25.17 & 76.00 & 90.07 & 46.47 & 60.60 & 62.92 & \cellcolor{blue!18}\textbf{53.04} \\
\midrule
\cellcolor{blue!18}\textbf{RotateKV 4-bit} & $4.00\times$ & 39.62 & 24.91 & 26.48 & 77.50 & 88.31 & 46.94 & 59.88 & 61.49 & \cellcolor{blue!18}\textbf{53.14} \\
TurboQuant K4V4 & $4.00\times$ & 37.77 & 25.59 & 25.73 & 76.00 & 88.23 & 46.25 & 58.69 & 61.29 & 52.44 \\
RotateKV 3-bit & $5.33\times$ & 34.46 & 25.75 & 26.63 & 76.00 & 87.90 & 45.93 & 59.98 & 61.96 & 52.33 \\
TurboQuant K4V2 & $5.33\times$ & 35.64 & 24.59 & 26.14 & 76.50 & 89.53 & 46.36 & 56.38 & 59.61 & 51.84 \\
\cellcolor{blue!7}\underline{BellowsQ $n{=}3$ INT8} & $5.46\times$ & 39.24 & 24.92 & 25.63 & 76.00 & 89.57 & 47.40 & 59.98 & 61.60 & \cellcolor{blue!7}\underline{53.04} \\
\midrule
\cellcolor{blue!18}\textbf{BellowsQ $n{=}2$ INT4} & $6.81\times$ & 37.45 & 25.56 & 25.65 & 76.00 & 88.56 & 46.84 & 59.73 & 62.26 & \cellcolor{blue!18}\textbf{52.76} \\
KIVI K2V2 & $7.11\times$ & 36.94 & 24.43 & 25.86 & 76.00 & 89.27 & 46.83 & 58.75 & 58.96 & 52.13 \\
OTT K2V2 & $7.11\times$ & 36.55 & 24.66 & 26.36 & 76.00 & 89.47 & 47.03 & 60.25 & 59.79 & 52.51 \\
RotateKV 2-bit & $8.00\times$ & 31.06 & 24.24 & 26.04 & 72.50 & 85.35 & 42.75 & 55.79 & 54.37 & 49.01 \\
TurboQuant K2V2 & $8.00\times$ & 34.83 & 23.12 & 25.06 & 71.50 & 85.96 & 44.64 & 50.24 & 51.13 & 48.31 \\
\cellcolor{blue!7}\underline{BellowsQ $n{=}2$ INT3} & $8.63\times$ & 39.19 & 25.30 & 26.05 & 76.00 & 88.72 & 45.93 & 59.22 & 60.99 & \cellcolor{blue!7}\underline{52.67} \\
\midrule
\cellcolor{blue!7}\underline{BellowsQ $n{=}4$ INT4} & $12.98\times$ & 38.27 & 25.56 & 24.81 & 75.50 & 89.57 & 46.67 & 60.09 & 62.13 & \cellcolor{blue!7}\underline{52.83} \\
BellowsQ $n{=}4$ INT3 & $16.25\times$ & 35.95 & 25.49 & 25.34 & 76.00 & 90.32 & 47.03 & 60.03 & 61.88 & 52.75 \\
\cellcolor{blue!18}\textbf{BellowsQ $n{=}4$ K3V2} & $18.59\times$ & 37.83 & 25.23 & 24.55 & 76.00 & 89.82 & 46.88 & 60.47 & 62.21 & \cellcolor{blue!18}\textbf{52.87} \\
\bottomrule
\end{tabular}
}
\vspace{3pt}
\end{table}

\begin{table}[H]
\centering
\caption{\textbf{Protocol A, Qwen2.5-7B.} Absolute LongBench scores, full test sets, every BellowsKV point and every official baseline configuration, grouped by measured compression tier (rules separate tiers; bins in Table~\ref{app:tab:tiers}); within a tier rows are sorted by ratio and the \colorbox{blue!18}{\textbf{best}} and \colorbox{blue!7}{\underline{second}} Avg are marked. Baselines run each method's official code; PolarQuant runs its Mixed K+V path, and on Qwen it and TurboQuant quantize bias-free keys (Appendix~\ref{app:exp_setup}).}
\label{app:tab:lb_qwen257b}
\vspace{5pt}
\fontsize{9pt}{11pt}\selectfont
\renewcommand{\arraystretch}{1.0}
\setlength{\tabcolsep}{5pt}
\resizebox{0.98\linewidth}{!}{%
\begin{tabular}{@{}lrccccccccc@{}}
\toprule
Method & Comp. & Qasper & QMSum & MNews & TREC & TQA & SAMSum & LCC & RB-P & Avg \\
\midrule
Dense fp16 & $1.00\times$ & 43.52 & 23.61 & 23.98 & 71.50 & 89.64 & 45.98 & 60.77 & 66.82 & \textbf{53.23} \\
\midrule
BellowsS $n{=}2$ & $1.92\times$ & 42.82 & 24.16 & 23.68 & 71.00 & 88.67 & 45.05 & 56.66 & 63.01 & 51.88 \\
BellowsS $n{=}3$ & $2.86\times$ & 42.99 & 23.50 & 23.48 & 71.00 & 89.00 & 45.12 & 58.00 & 62.25 & 51.92 \\
\cellcolor{blue!7}\underline{BellowsQ $n{=}2$ INT8} & $3.57\times$ & 43.22 & 24.06 & 23.58 & 71.00 & 88.60 & 44.91 & 57.27 & 63.34 & \cellcolor{blue!7}\underline{52.00} \\
\cellcolor{blue!18}\textbf{PolarQuant r4t4v4} & $3.66\times$ & 43.94 & 24.42 & 25.52 & 72.00 & 89.46 & 45.76 & 63.68 & 66.41 & \cellcolor{blue!18}\textbf{53.90} \\
BellowsS $n{=}4$ & $3.78\times$ & 42.20 & 23.80 & 22.94 & 71.00 & 88.15 & 44.65 & 55.44 & 61.51 & 51.21 \\
\midrule
\cellcolor{blue!7}\underline{TurboQuant K4V4} & $4.00\times$ & 42.65 & 23.53 & 23.89 & 72.00 & 88.38 & 45.72 & 61.34 & 66.33 & \cellcolor{blue!7}\underline{52.98} \\
\cellcolor{blue!18}\textbf{PolarQuant r3t3v4} & $4.13\times$ & 44.13 & 24.02 & 25.54 & 72.00 & 90.10 & 46.49 & 64.07 & 64.75 & \cellcolor{blue!18}\textbf{53.89} \\
PolarQuant r2t2v4 & $4.74\times$ & 38.51 & 22.70 & 24.87 & 69.00 & 89.54 & 44.25 & 60.61 & 57.45 & 50.87 \\
BellowsQ $n{=}3$ INT8 & $5.28\times$ & 43.01 & 23.87 & 23.43 & 71.00 & 88.34 & 45.47 & 58.31 & 61.75 & 51.90 \\
TurboQuant K4V2 & $5.33\times$ & 42.56 & 23.27 & 23.53 & 72.00 & 88.29 & 45.30 & 55.14 & 59.88 & 51.25 \\
BellowsQ $n{=}2$ INT4 & $6.41\times$ & 42.05 & 24.13 & 23.48 & 71.50 & 88.91 & 45.48 & 56.31 & 63.05 & 51.86 \\
\midrule
\cellcolor{blue!7}\underline{PolarQuant r2t2v2} & $6.74\times$ & 37.63 & 23.09 & 25.41 & 67.50 & 88.08 & 44.44 & 58.11 & 56.74 & \cellcolor{blue!7}\underline{50.12} \\
TurboQuant K2V2 & $8.00\times$ & 36.18 & 21.85 & 22.25 & 66.50 & 81.39 & 42.92 & 55.44 & 50.93 & 47.18 \\
\cellcolor{blue!18}\textbf{BellowsQ $n{=}2$ INT3} & $8.01\times$ & 41.80 & 23.37 & 23.59 & 70.50 & 89.26 & 44.97 & 57.24 & 62.09 & \cellcolor{blue!18}\textbf{51.60} \\
\midrule
\cellcolor{blue!18}\textbf{BellowsQ $n{=}4$ INT4} & $12.15\times$ & 41.84 & 23.54 & 22.60 & 71.00 & 88.60 & 45.45 & 55.98 & 59.98 & \cellcolor{blue!18}\textbf{51.12} \\
BellowsQ $n{=}4$ INT3 & $14.96\times$ & 41.21 & 23.85 & 22.17 & 70.50 & 87.21 & 44.76 & 53.26 & 57.18 & 50.02 \\
\cellcolor{blue!7}\underline{BellowsQ $n{=}4$ K3V2} & $16.92\times$ & 40.13 & 24.44 & 23.84 & 71.00 & 86.45 & 45.74 & 53.27 & 57.40 & \cellcolor{blue!7}\underline{50.28} \\
\bottomrule
\end{tabular}
}
\vspace{3pt}
\end{table}

\subsection{Protocol C: full leaderboards}
\label{app:protoC_full}
\begin{table}[ht]
\centering
\footnotesize
\caption{\textbf{Protocol~C Full Table} Single-turn retrieve-then-read on BrowseComp-Plus: the
benchmark's oracle prompt filled to $32{,}768$ tokens with its own mined hard negatives, all $830$
queries, greedy, the identical context for every method so dense is the ceiling. Score is the
benchmark's own judge and Comp.\ is measured compression with all metadata counted.}
\label{app:tab:protoc}
\vspace{3pt}
\setlength{\tabcolsep}{6pt}
\begin{tabular}{@{}lrr@{}}
\toprule
Method & Comp. & Score \\
\midrule
\multicolumn{3}{@{}l}{\emph{Qwen3-32B}} \\
\midrule
Dense fp16 & $1.00\times$ & 85.54 \\
PolarQuant$^\dagger$ & $3.66\times$ & 83.49 \\
KIVI$^\dagger$ & $3.76\times$ & 84.94 \\
xKV$^\dagger$ & $4.00\times$ & 78.07 \\
PolarQuant$^\dagger$ & $6.74\times$ & 51.93 \\
OTT$^\dagger$ & $7.11\times$ & 77.95 \\
KIVI$^\dagger$ & $7.11\times$ & 75.42 \\
\textbf{BellowsQ} & $7.42\times$ & \textbf{84.22} \\
CommonKV$^\dagger$ & $7.96\times$ & 81.08 \\
xKV$^\dagger$ & $8.00\times$ & 26.51 \\
SPECTRA & $9.31\times$ & 81.45 \\
SPECTRA & $11.90\times$ & 78.92 \\
\textbf{BellowsQ} & $14.61\times$ & \textbf{86.02} \\
CommonKV$^\dagger$ & $15.91\times$ & 66.02 \\
xKV$^\dagger$ & $16.00\times$ & \phantom{0}8.43 \\
\midrule
\multicolumn{3}{@{}l}{\emph{Qwen3-30B-A3B-Instruct-2507}} \\
\midrule
Dense fp16 & $1.00\times$ & 86.14 \\
xKV$^\dagger$ & $2.00\times$ & 82.77 \\
PolarQuant$^\dagger$ & $3.66\times$ & 80.84 \\
xKV$^\dagger$ & $4.00\times$ & 15.54 \\
SPECTRA & $5.36\times$ & 85.66 \\
SPECTRA & $5.58\times$ & 83.13 \\
PolarQuant$^\dagger$ & $6.74\times$ & \phantom{0}4.22 \\
OTT$^\dagger$ & $7.11\times$ & 81.45 \\
KIVI$^\dagger$ & $7.11\times$ & 77.59 \\
\textbf{BellowsQ} & $7.42\times$ & \textbf{85.30} \\
CommonKV$^\dagger$ & $7.96\times$ & 49.88 \\
xKV$^\dagger$ & $8.00\times$ & \phantom{0}5.30 \\
\textbf{BellowsQ} & $14.61\times$ & \textbf{85.78} \\
\bottomrule
\end{tabular}
\end{table}

\subsection{The eviction grid in full}
\label{app:evict_full}

Table~\ref{app:tab:evict_tiers} is the complete grid behind Section~\ref{sec:exp_evict}, all
three models and every budget tier from $\sim$2$\times$ to $\sim$57$\times$, each cell with its
measured compression. Rows in one tier are the equal-memory pairs; no pair is byte-exact,
BellowsE stores $1$--$5\%$ more than the control at the paired points, and the margins are
several times that gap. The $\sim$2$\times$ and $\sim$4$\times$ tiers hold only the control
(BellowsE's shallowest point is already $\sim$8$\times$), and the same-keep-ratio comparison,
where the control wins by $0.8$--$3.3$ points while storing $3.8\times$ more, is read across
tiers: the BellowsE row of one tier against the control row two tiers below.
Table~\ref{app:tab:exchange} prices the two currencies directly: from either keep ratio,
pruning four times harder buys $\sim$4$\times$ for $2.6$--$7.8$ points, while sharing across
four layers buys $\sim$3.9$\times$ for $0.8$--$1.7$ points.

\begin{table}[H]
\centering
\caption{\textbf{The eviction axis by budget tier (Protocol B, full sample; retention \%).} Every route runs the identical verdict; rows are grouped by measured compression so that each tier compares like with like: the per-layer control prunes its own fp16 cache, BellowsE prunes the shared latent, and at $\sim$30$\times$ the composed point meets the naive stack and deep eviction. Comp. spans the three models (Qwen's KV width differs). Best per tier in bold.}
\label{app:tab:evict_tiers}
\vspace{5pt}
\small
\renewcommand{\arraystretch}{1.09}
\begin{tabular}{@{}lllcccc@{}}
\toprule
Budget & Route & Keep & Comp. & Llama & Mistral & Qwen \\
\midrule
\multirow{1}{*}{$\sim$2$\times$} & Per-layer eviction & 1/2 & 2.0 & 98.4 & 99.4 & 100.9 \\
\midrule
\multirow{1}{*}{$\sim$4$\times$} & Per-layer eviction & 1/4 & 4.0 & 97.3 & 97.7 & 99.5 \\
\midrule
\multirow{2}{*}{$\sim$8$\times$} & BellowsE & 1/2 & 7.7--7.8 & \textbf{97.6} & \textbf{97.9} & \textbf{99.2} \\
 & Per-layer eviction & 1/8 & 8.0 & 94.7 & 94.9 & 98.3 \\
\midrule
\multirow{2}{*}{$\sim$16$\times$} & BellowsE & 1/4 & 15.2--15.4 & \textbf{96.1} & \textbf{96.4} & \textbf{97.9} \\
 & Per-layer eviction & 1/16 & 16.0 & 89.5 & 90.7 & 92.6 \\
\midrule
\multirow{3}{*}{$\sim$30$\times$} & BellowsQ$\times$BellowsE & 1/2 & 26.3--27.7 & \textbf{97.5} & 97.9 & \textbf{99.2} \\
 & BellowsE & 1/8 & 29.6--30.0 & 92.5 & 93.8 & 95.0 \\
 & xKV$+$INT4$+$pooled SnapKV & 1/2 & 30.1 & 96.1 & \textbf{98.4} & 23.8 \\
\midrule
\multirow{1}{*}{$\sim$57$\times$} & BellowsE & 1/16 & 56.6--57.4 & 87.3 & 87.6 & 90.1 \\
\bottomrule
\end{tabular}
\vspace{3pt}
\end{table}

\begin{table}[H]
\centering
\caption{\textbf{The exchange rate between the two currencies.} Retention points
paid to buy $\sim$4$\times$ by pruning four times harder, against points paid to
buy $\sim$3.9$\times$ by sharing across four layers at a fixed keep ratio.}
\label{app:tab:exchange}
\vspace{7pt}
\small
\renewcommand{\arraystretch}{1.13}
\begin{tabular}{lccc}
\toprule
From keep & Model & prune harder & share \\
\midrule
1/2 & Llama-3.1-8B & 3.74 & 0.82 \\
 & Mistral-7B & 4.47 & 1.45 \\
 & Qwen2.5-7B & 2.63 & 1.74 \\
\addlinespace[3.5pt]
1/4 & Llama-3.1-8B & 7.80 & 1.24 \\
 & Mistral-7B & 7.02 & 1.28 \\
 & Qwen2.5-7B & 6.87 & 1.57 \\
\bottomrule
\end{tabular}
\vspace{3pt}
\end{table}

Table~\ref{app:tab:compose} is the $\sim$30$\times$ class of Section~\ref{sec:exp_evict} on all three
models: the composed operating point against the naive stack and the deep-eviction arm.

\begin{table}[H]
\centering
\caption{\textbf{The $\sim$30$\times$ budget class, all three models} (Protocol B,
full sample; retention \%). The composed point stores slightly more than the stack's
nominal ratio (27.7/26.3$\times$ against 30.1$\times$, a 9--13\% byte gap in its
favour); the stack's verdict is pooled per group because per-layer verdicts cannot
share a basis. Setup in Appendix~\ref{app:exp_setup}.}
\label{app:tab:compose}
\vspace{7pt}
\small
\renewcommand{\arraystretch}{1.13}
\begin{tabular}{lcccc}
\toprule
Route & Comp. & Llama & Mistral & Qwen \\
\midrule
BellowsQ$\times$BellowsE (evict 1/2) & 27.7/26.3 & \textbf{97.5} & 97.9 & \textbf{99.2} \\
naive stack: xKV$+$INT4$+$pooled SnapKV & 30.1 & 96.1 & \textbf{98.4} & 23.8 \\
BellowsE alone (fp16, evict 1/8) & 30.0/29.6 & 92.5 & 93.8 & 95.0 \\
\bottomrule
\end{tabular}
\vspace{3pt}
\end{table}

\subsection{Equal bit-width on the quantization axis}
\label{app:quant_full}

The tiered tables of Appendix~\ref{app:protoA_full} are the equal-memory comparison: rows in
one tier store the same bytes to within the tier's width. Table~\ref{app:tab:eqbw} asks the
complementary question by holding the bit-width fixed and varying only what is quantized, each
layer's own cache or one latent shared by $n$ layers, which prices the sharing factor itself at
equal bits.

\begin{table}[H]
\centering
\caption{\textbf{Equal bit-width (Protocol A, full sample).} At a fixed bit-width
the only difference is what is quantized: each layer's own cache, or one latent
shared by $n$ layers. LongBench average and measured compression. Qwen's per-layer
baselines are PolarQuant and TurboQuant, both with bias-free keys (Appendix~\ref{app:exp_setup}).}
\label{app:tab:eqbw}
\vspace{7pt}
\small
\renewcommand{\arraystretch}{1.13}
\begin{tabular}{llccc}
\toprule
Bits & Method & Comp. & Avg & vs best baseline \\
\midrule
\multicolumn{5}{c}{\textit{Llama-3.1-8B-Instruct  (dense 53.29)}} \\
\midrule
INT4 & KIVI & 3.76 & 53.92 & -- \\
 & OTT & 3.76 & 53.83 & -- \\
 & PolarQuant & 3.66 & 53.66 & -- \\
 & TurboQuant & 4.00 & 52.84 & -- \\
 & BellowsQ $n{=}2$ & 6.75 & 52.92 & -1.00 at 1.8$\times$ its compression \\
 & BellowsQ $n{=}4$ & 12.73 & 52.45 & -1.47 at 3.4$\times$ its compression \\
\addlinespace[2.5pt]
INT3 & RotateKV & 5.33 & 52.38 & -- \\
 & BellowsQ $n{=}2$ & 8.53 & 52.72 & +0.34 at 1.6$\times$ its compression \\
 & BellowsQ $n{=}4$ & 15.85 & 52.42 & +0.04 at 3.0$\times$ its compression \\
\midrule
\multicolumn{5}{c}{\textit{Mistral-7B-Instruct-v0.3  (dense 52.78)}} \\
\midrule
INT4 & RotateKV & 4.00 & 53.14 & -- \\
 & OTT & 3.76 & 52.99 & -- \\
 & KIVI & 3.76 & 52.80 & -- \\
 & TurboQuant & 4.00 & 52.44 & -- \\
 & BellowsQ $n{=}2$ & 6.81 & 52.76 & -0.38 at 1.7$\times$ its compression \\
 & BellowsQ $n{=}4$ & 12.98 & 52.83 & -0.31 at 3.2$\times$ its compression \\
\addlinespace[2.5pt]
INT3 & RotateKV & 5.33 & 52.33 & -- \\
 & BellowsQ $n{=}2$ & 8.63 & 52.67 & +0.34 at 1.6$\times$ its compression \\
 & BellowsQ $n{=}4$ & 16.25 & 52.75 & +0.42 at 3.0$\times$ its compression \\
\midrule
\multicolumn{5}{c}{\textit{Qwen2.5-7B-Instruct  (dense 53.23)}} \\
\midrule
INT4 & PolarQuant & 3.66 & 53.90 & -- \\
 & TurboQuant & 4.00 & 52.98 & -- \\
 & BellowsQ $n{=}2$ & 6.41 & 51.86 & -2.04 at 1.8$\times$ its compression \\
 & BellowsQ $n{=}4$ & 12.15 & 51.12 & -2.78 at 3.3$\times$ its compression \\
\addlinespace[2.5pt]
INT3 & BellowsQ $n{=}2$ & 8.01 & 51.60 & no valid baseline at this bit-width \\
 & BellowsQ $n{=}4$ & 14.96 & 50.02 & no valid baseline at this bit-width \\
\bottomrule
\end{tabular}
\vspace{3pt}
\end{table}

\subsection{Adaptivity and whisper, per task}
\label{app:ablation_full}

\begin{figure}[H]
  \centering
  \includegraphics[width=0.5\linewidth]{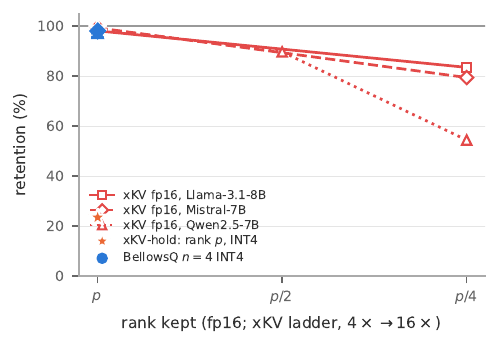}
  \caption{\textbf{The rank cliff on the rival's own ladder.} Protocol-B retention of xKV's fp16 basis
  against the rank kept ($p$, $p/2$, $p/4$, i.e. $4\times$, $8\times$, $16\times$), three models, with
  xKV-hold (the same basis at rank $p$ stored at INT4, $15.1\times$) and BellowsQ $n{=}4$ INT4 at full rank.
  Cutting rank falls; cutting bits at full rank does not, except on Qwen, where every arm that stores
  xKV's basis at four bits collapses (Section~\ref{sec:exp_rank}).}
  \label{fig:rankcliff}
\end{figure}

Table~\ref{tab:rank} in the main text holds the rank ladder and the transplant of Section~\ref{sec:exp_rank},
Table~\ref{app:tab:transplant} the transplant per task on all three models, and
Figure~\ref{fig:rankcliff} draws xKV's rank ladder against the rank kept.
Table~\ref{app:tab:ablation} holds the adaptivity and whisper rows behind
Section~\ref{sec:exp_rank} per task. The whisper reading is in
Appendix~\ref{app:whisper}: its average effect is small ($\pm0.1$ at the high tier), its value
is the worst case (the 2WikiMQA column), and the V-whisper row's gain at 32k flips sign on
short-prefill inputs, which is why it is not the method.

\begin{table}[H]
\centering
\caption{\textbf{The transplant, per task.} Protocol B retention (\%) of xKV at $16\times$ (fp16, rank $p/4$), xKV-hold (the same per-sequence basis at near-full rank with our INT4 leg, $15.1\times$) and BellowsQ $n{=}4$ INT4, all three models. The recovery lands on the tasks truncation had cost; on Qwen the arm collapses on every task.}
\label{app:tab:transplant}
\vspace{5pt}
\fontsize{9pt}{11pt}\selectfont
\renewcommand{\arraystretch}{1.0}
\setlength{\tabcolsep}{5pt}
\resizebox{0.98\linewidth}{!}{%
\begin{tabular}{@{}lrcccccccccccc@{}}
\toprule
Arm & Comp. & Qsp & MFQ & HQA & MSQ & NQA & Gov & QMS & PsR & LCC & RBP & 2Wk & Avg \\
\specialrule{\lightrulewidth}{2.5pt}{3.5pt}
\multicolumn{14}{c}{\fontsize{10.5pt}{11pt}\selectfont\textbf{Llama-3.1-8B-Instruct}} \\
\addlinespace[3pt]
\midrule
xKV (fp16, rank $p/4$) & $16.00\times$ & 91.4 & 80.9 & 82.4 & 87.3 & 73.3 & 75.6 & 94.4 & 95.5 & 92.8 & 86.8 & 58.5 & 83.5 \\
xKV-hold & $15.06\times$ & 97.5 & 95.4 & 96.0 & 100.9 & 96.5 & 96.4 & 99.2 & 99.5 & 101.6 & 97.4 & 92.9 & 97.6 \\
BellowsQ $n{=}4$ INT4 & $13.84\times$ & 94.6 & 99.0 & 99.1 & 96.4 & 96.2 & 93.5 & 98.7 & 100.0 & 99.7 & 99.0 & 101.5 & 98.0 \\
\specialrule{\heavyrulewidth}{5pt}{1.4pt}
\specialrule{\lightrulewidth}{0pt}{3.5pt}
\multicolumn{14}{c}{\fontsize{10.5pt}{11pt}\selectfont\textbf{Mistral-7B-Instruct-v0.3}} \\
\addlinespace[3pt]
\midrule
xKV (fp16, rank $p/4$) & $16.00\times$ & 90.4 & 88.5 & 47.8 & 42.1 & 85.2 & 93.0 & 91.8 & 81.7 & 92.7 & 90.0 & 71.1 & 79.5 \\
xKV-hold & $15.06\times$ & 99.8 & 103.0 & 99.9 & 101.5 & 93.7 & 99.4 & 96.9 & 94.9 & 99.8 & 98.9 & 102.9 & 99.2 \\
BellowsQ $n{=}4$ INT4 & $13.84\times$ & 93.9 & 103.1 & 99.1 & 97.2 & 96.3 & 93.6 & 99.8 & 97.4 & 98.6 & 101.0 & 99.3 & 98.1 \\
\specialrule{\heavyrulewidth}{5pt}{1.4pt}
\specialrule{\lightrulewidth}{0pt}{3.5pt}
\multicolumn{14}{c}{\fontsize{10.5pt}{11pt}\selectfont\textbf{Qwen2.5-7B-Instruct}} \\
\addlinespace[3pt]
\midrule
xKV (fp16, rank $p/4$) & $16.00\times$ & 63.0 & 69.6 & 26.1 & 22.2 & 38.2 & 80.8 & 78.3 & 48.5 & 74.7 & 61.5 & 35.9 & 54.4 \\
xKV-hold & $15.06\times$ & 17.3 & 16.5 & 26.9 & 28.3 & 22.3 & 10.2 & 31.6 & 30.2 & 21.1 & 29.2 & 24.1 & 23.4 \\
BellowsQ $n{=}4$ INT4 & $13.13\times$ & 100.8 & 97.3 & 98.6 & 99.2 & 96.8 & 93.6 & 99.9 & 99.0 & 92.6 & 94.5 & 98.6 & 97.3 \\
\bottomrule
\end{tabular}
}
\vspace{3pt}
\end{table}

\begin{table}[H]
\centering
\caption{\textbf{Adapting to the sequence buys nothing the frozen consensus lacks, and the whisper's value is
the worst case, not the mean.} Adaptivity and whisper ablation, per task (Llama-3.1-8B, Protocol B, full
sample), the rows behind Section~\ref{sec:exp_rank}; the oracle re-decides the basis per sequence at
2.2$\times$ prefill cost. Setup in Appendix~\ref{app:exp_setup}.}
\label{app:tab:ablation}
\vspace{5pt}
\fontsize{9pt}{11pt}\selectfont
\renewcommand{\arraystretch}{1.0}
\setlength{\tabcolsep}{5pt}
\resizebox{0.98\linewidth}{!}{%
\begin{tabular}{@{}lrcccccccccccc@{}}
\toprule
Variant & Comp. & Qsp & MFQ & HQA & MSQ & NQA & Gov & QMS & PsR & LCC & RBP & 2Wk & Avg \\
\midrule
BellowsQ (final) & 13.84 & 94.6 & 99.0 & 99.1 & 96.4 & 96.2 & 93.5 & 98.7 & 100.0 & 99.7 & 99.0 & 101.5 & 98.0 \\
\quad + V-whisper & 13.13 & 94.9 & 99.1 & 98.2 & 98.5 & 99.3 & 95.2 & 100.7 & 99.5 & 101.0 & 98.9 & 100.7 & 98.7 \\
\quad $-$ whisper & 13.90 & 93.3 & 97.5 & 99.0 & 97.1 & 97.5 & 95.4 & 100.4 & 100.0 & 100.7 & 98.6 & 97.5 & 97.9 \\
\quad + per-sequence gauge & 13.90 & 93.2 & 94.8 & 99.5 & 98.5 & 93.5 & 92.3 & 98.6 & 99.0 & 99.1 & 97.8 & 97.8 & 96.7 \\
per-seq SVD oracle & 15.06 & 91.6 & 92.8 & 98.8 & 99.5 & 95.3 & 99.9 & 99.5 & 99.5 & 96.3 & 96.4 & 101.2 & 97.3 \\
xKV-hold (+ our INT4) & 15.06 & 97.5 & 95.4 & 96.0 & 100.9 & 96.5 & 96.4 & 99.2 & 99.5 & 101.6 & 97.4 & 92.9 & 97.6 \\
8$\times$ point ($n{=}2$, INT4) & 6.90 & 96.3 & 97.7 & 100.3 & 97.6 & 98.6 & 98.6 & 101.1 & 100.0 & 99.6 & 99.2 & 101.6 & 99.2 \\
12$\times$ point ($n{=}3$, INT4) & 10.40 & 93.8 & 96.7 & 99.1 & 102.5 & 97.2 & 97.0 & 99.9 & 100.0 & 99.4 & 98.9 & 100.3 & 98.6 \\
\bottomrule
\end{tabular}
}
\vspace{3pt}
\end{table}

\end{document}